\documentclass[10pt,twocolumn,letterpaper]{article}

\usepackage[pagenumbers]{cvpr} 

\definecolor{cvprblue}{rgb}{0.21,0.49,0.74}
\usepackage[pagebackref,breaklinks,colorlinks,allcolors=cvprblue]{hyperref}
\usepackage{multirow}
\usepackage{colortbl}
\usepackage{booktabs}
\usepackage{array}
\usepackage[table]{xcolor}
\usepackage{cuted}
\usepackage{algorithm}
\usepackage{algpseudocode}
\usepackage{fontawesome5}
\newcommand{\venue}[2]{{\scriptsize (#1)~\cite{#2}}}

\def\confName{CVPR}
\def\confYear{2027}

\makeatletter

\patchcmd{\@maketitle}
  {\vspace*{12pt}}
  {\vspace*{4pt}}
  {}{}

\patchcmd{\@maketitle}
  {\vspace*{24pt}}
  {\vspace*{12pt}}
  {}{}

\makeatother

\title{
\mbox{%
    \makebox[0pt][r]{%
        \raisebox{-0.3\height}{%
            \includegraphics[height=3.2em]{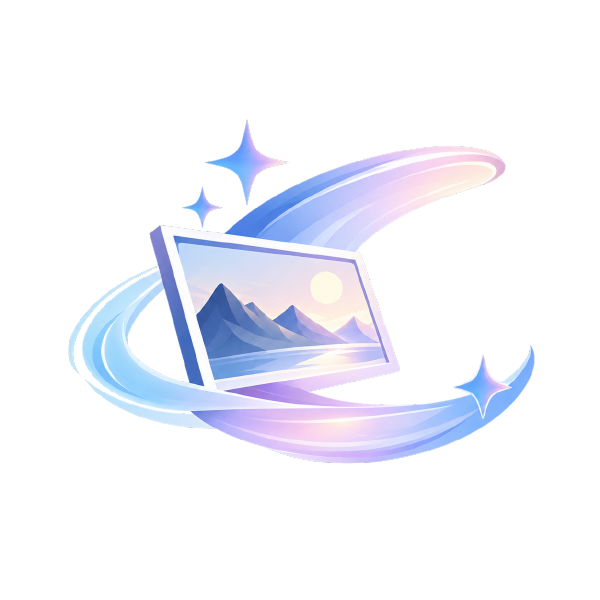}%
        }%
        \hspace{0.4em}%
    }%
    Transfer the Style, Not the Semantics:%
}\\
Mitigating Semantic Leakage in Image-Guided Stylization
}

\author{
Haojun Tang$^{1,2,*,\dagger}$ \hspace{0.3em}
Qiwei Lin$^{3,*}$ \hspace{0.3em}
Tongda Xu$^{1,*}$ \hspace{0.3em} \\
Wenda Zhao$^{2,}$\textsuperscript{\faEnvelope} \hspace{0.3em}
Yian Zhao$^{4}$ \hspace{0.3em}
Yu Zhang$^{2}$ \hspace{0.3em}
Lida Huang$^{1}$ \hspace{0.3em}
Yan Wang$^{1,}$\textsuperscript{\faEnvelope}
\\[0.5em]
$^{1}$Institute for AI Industry Research (AIR), Tsinghua University \\
$^{2}$School of Information and Communication Engineering, Dalian University of Technology \\
$^{3}$Beijing Institute of Radio Measurement \\
$^{4}$School of Information Engineering, Peking University \\
{\tt\small tanghaojun@mail.dlut.edu.cn \hspace{0.3em}
zhaowenda@dlut.edu.cn \hspace{0.3em} wangyan@air.tsinghua.edu.cn}
}

\begin{document}
\twocolumn[{%
\renewcommand\twocolumn[1][]{#1}%
\maketitle
\begin{center}
    \centering
    \includegraphics[width=\textwidth]{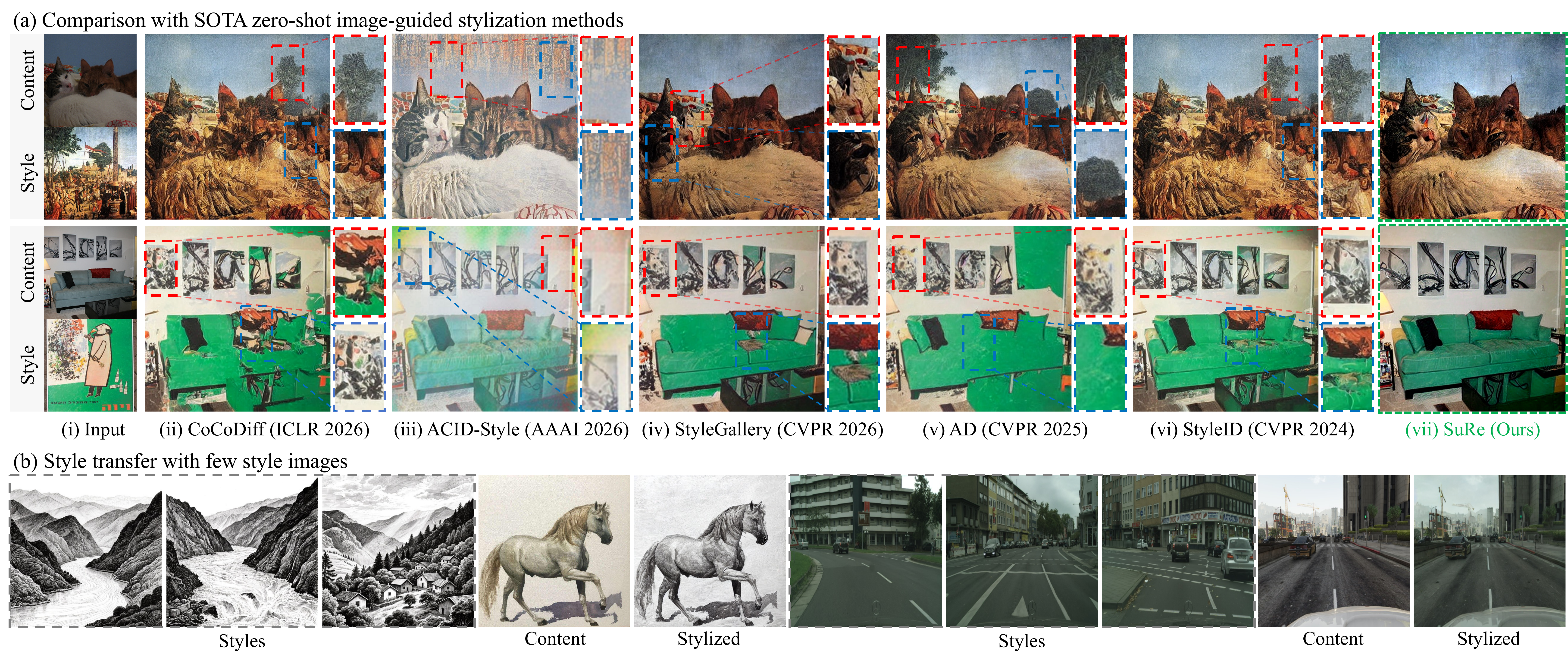}
    \captionof{figure}{Teaser of SuRe. (a) Compared with SOTA zero-shot image-guided stylization methods, SuRe better mitigates semantic leakage while preserving content structure and target style. (b) SuRe naturally extends to stylization with few style reference images.}
    \label{Teaser}
\end{center}%
}]

\begingroup
\makeatletter
\renewcommand{\@makefntext}[1]{\noindent #1}
\makeatother
\renewcommand{\thefootnote}{}
\footnotetext{%
    \begin{tabular}[t]{@{}l@{}}
        $*$ Equal contribution. \quad
        \faEnvelope\hspace{0.2em} Corresponding authors. \\[2pt]
        $\dagger$ Work done by Haojun Tang as a student researcher at AIR.
    \end{tabular}
}
\endgroup



\begin{abstract}
Image-guided zero-shot style transfer enables training-free stylization of arbitrary content images using a reference style.
Despite recent progress, existing methods commonly rely on self-attention manipulation to inject style information, which may transfer semantic content from the style image, leading to semantic leakage.
In this work, we revisit this phenomenon from the perspective of key--value coupling in the self-attention of diffusion models.
Our analysis shows that the native correspondence between style keys and values plays an important role in transferring semantic information from the style image.
Motivated by this observation, we propose SuRe, a timestep-aware key--value regulation framework that mitigates semantic leakage by disrupting the native key--value correspondence while progressively restoring it for faithful style transfer.
Specifically, we present value shuffle to decouple style keys and values while preserving the empirical distribution of the values, together with a progressive recoupling mechanism that gradually recouples them throughout the stylization process.
Experiments against 18 representative methods demonstrate that SuRe maintains competitive style similarity while consistently improving content similarity across SD1.5, SDXL, SD3, and FLUX.1-dev, with natural generalization to multi-style transfer.
The code will be publicly available at
\href{https://xinr-tang.github.io/SuRe-homepage/}{\textcolor{cvprblue}{https://xinr-tang.github.io/SuRe-homepage/}}.
\end{abstract}
\vspace{-6pt}

\begin{figure}[t]
    \centering
    \includegraphics[width=\linewidth]{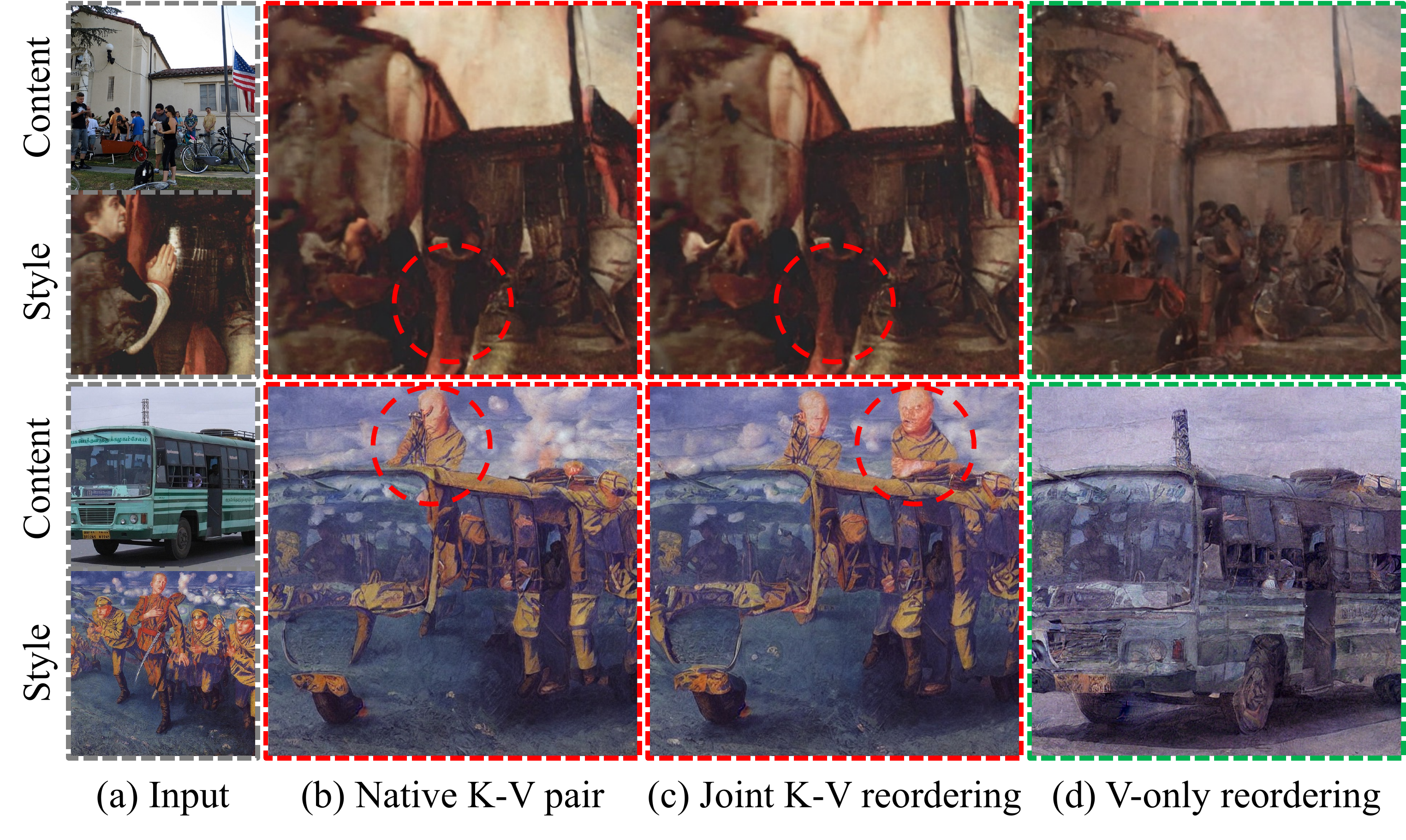}
    \caption{Qualitative comparison of different key--value configurations. Preserving the K--V pairing leads to semantic leakage, while value-only reordering suppresses leaked semantics.}
    \label{fig:visual}
\end{figure}

\section{Introduction}
\label{sec:intro}
Zero-shot style transfer aims to stylize a content image according to a reference style image while preserving its semantic content, without training a task-specific stylization model.
Recent studies~\cite{cocodiff,ACID-Style,StyleID,Stylekeeper,Z*} address this task by leveraging the powerful generative priors of pretrained diffusion models, achieving better stylization results than previous CNN-based~\cite{AdaIN, MAST, EFDM, CAST, Aespa-net} and Transformer-based methods~\cite{AdaAttN, StyTR2, Puff-net}. 
Among these methods, self-attention manipulation has become a key technique. 
Specifically, the core idea is to combine content queries with style keys and values in self-attention layers, thereby transferring stylistic cues from the style image to the content image.
Despite its effectiveness, such an approach often introduces unintended semantic information from the style image into the stylized output, resulting in semantic leakage, as illustrated in Fig.~\ref{Teaser}.

To mitigate this phenomenon, some recent methods explore feature masking~\cite{zhu2025less}, feature purification~\cite{CleanStyle}, negative guidance~\cite{Stylekeeper}, or patch suppression~\cite{Only-Style}. 
Although these methods can alleviate semantic leakage, they are primarily designed for text-to-image stylization, where the semantic content specified by text prompts is relatively abstract and does not impose a fixed spatial layout.
In image-guided stylization, recent methods~\cite{Stylessp,StyleGallery} mitigate style semantics by constraining the sampling process or style matching to preserve the semantic structure of the content image.
However, the underlying reason why semantic information from the style image is transferred remains unexplored.

\begin{figure}[t]
    \centering
    \includegraphics[width=\linewidth]{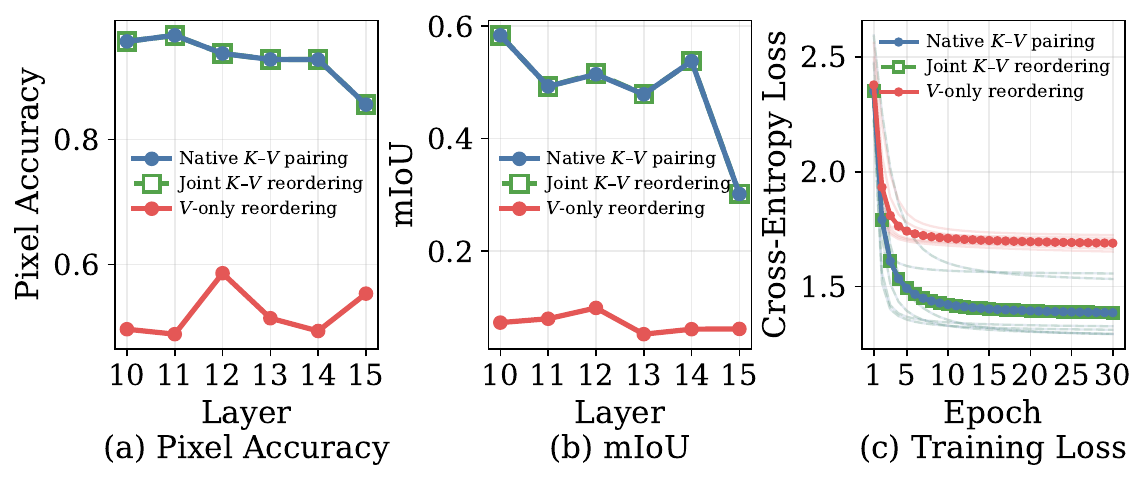}
    \caption{Linear probing of different key--value configurations. Preserving the native K--V pairing retains the semantics of the style image, while value-only reordering reduces it.}
    \label{fig:LR}
\end{figure}

In this work, we revisit semantic leakage from the perspective of key--value coupling in the self-attention of diffusion models.
For $i$-th content query $Q_c^i$, the style key $K_s^j$ determines how strongly the corresponding value $V_s^j$ is attended to, while $V_s^j$ determines what information is transferred to location $i$.
Since each style key is inherently paired with its value at the same spatial location, if $Q_c^i$ and $K_s^j$ exhibit a high attention score, $V_s^j$ will contribute strongly to the output.
Consequently, both the stylistic and semantic information encoded in $V_s^j$ are transferred together, leading to semantic leakage.
For intuitive understanding, we conduct a simple experiment to validate the coupling relationship between keys and values.
Specifically, we compare three settings using attention distillation~\cite{AD}: (1) Native $K$--$V$ pairing, which uses the original $Q_c$, $K_s$, and $V_s$; (2) Joint $K$--$V$ reordering, which applies the same permutation matrix $P$ to $K_s$ and $V_s$, preserving their pairing while changing their spatial indices; and (3) $V$-only reordering, which applies $P$ only to $V_s$, thereby breaking the native key--value pairing.
As shown in Fig.~\ref{fig:visual}, both Native $K$--$V$ pairing and Joint $K$--$V$ reordering transfer semantic content from the style image, whereas $V$-only reordering suppresses semantic leakage.
To quantify this observation, we linearly probe the outputs of multiple attention layers under these settings on the Cityscapes~\cite{cordts2016cityscapesdatasetsemanticurban} and GTA5~\cite{GTA5} datasets.
Fig.~\ref{fig:LR} shows that Native $K$--$V$ pairing and Joint $K$--$V$ reordering achieve nearly identical pixel accuracy, mIoU, and convergence across layers. 
In contrast, $V$-only reordering substantially reduces pixel accuracy and mIoU while yielding a higher training loss.
These results reveal that breaking the key--value coupling reduces the semantic information of the style image encoded in the attention outputs.
Detailed explanations of this experiment and its mathematical analysis are provided in \textbf{Supp.}~\ref{sec:rationale}.
\textbf{These observations motivate a core question: how to break the coupling between style keys and values without sacrificing style fidelity?}

\begin{figure*}[t]
    \centering
    \includegraphics[width=\textwidth]{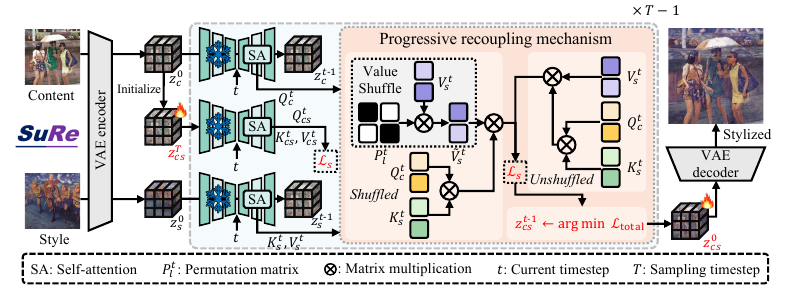}
    \caption{Overview of SuRe. The content and style images are first encoded into image latents. V-Shuffle permutes the style value tokens while keeping the style keys fixed, breaking their native correspondence while preserving the empirical value distribution. PRM then progressively combines shuffled and native attention during sampling, suppressing style semantics while refining the target style.}
    \label{fig:method}
\end{figure*}

To answer this question, we propose semantic uncoupling and recoupling (SuRe), a timestep-aware key--value regulation framework that suppresses semantic leakage for faithful image-guided stylization. 
Specifically, we present value shuffle (V-Shuffle), which randomly permutes style values while keeping the keys fixed, thereby breaking their original correspondence while preserving the empirical distribution of the value tokens.
Building upon V-Shuffle, we propose a progressive recoupling mechanism (PRM) that dynamically regulates key--value coupling during stylized-latent optimization.
PRM follows a cosine schedule strategy, emphasizing semantic uncoupling in early optimization steps and progressively restoring key--value coupling in later steps for fine-grained style transfer.
Experiments demonstrate that SuRe delivers superior overall performance against 18 representative methods in the standard single-style setting, remains effective across SD1.5, SDXL, SD3, and FLUX.1-dev, and naturally generalizes to multi-style transfer.
Our contributions are summarized as follows:
\begin{itemize}
    \item We revisit semantic leakage in image-guided zero-shot stylization from the perspective of key--value coupling and show that the coupling between style keys and values plays an important role in semantic leakage.

    \item We propose SuRe, a timestep-aware key--value regulation framework that combines V-Shuffle for semantic uncoupling with PRM for progressive recoupling, balancing semantic preservation and style fidelity.
    
    \item Experiments show that SuRe achieves superior overall performance over 18 comparison methods in the standard single-style setting across SD1.5, SDXL, SD3, and FLUX.1-dev, while naturally extending from single-style to multi-style transfer.
\end{itemize}

\section{Related Work}
\label{sec:related}

\subsection{Neural Style Transfer}
Neural style transfer synthesizes images that preserve content structure while adopting the appearance of a style image. 
Early CNN- and Transformer-based methods~\cite{Aespa-net, CAST, EFDM, MAST, AdaIN, StyTR2, AdaAttN} rely on feature-statistic matching, but are often limited by stylization capacity and loss design. 
Recent diffusion-based methods leverage generative priors through inversion~\cite{InST}, adapters~\cite{artadapter}, LoRA~\cite{Ziplora,K-lora}, or attention manipulation~\cite{Z*,StyleID,AD,ACID-Style}. 
In particular, attention-based methods directly regulate content--style interactions~\cite{StyleID,AD,ACID-Style,cocodiff,StyleGallery}, achieving strong stylization but often suffering from semantic leakage, where style-image semantics are transferred to the output. 
Our work aims to mitigate this issue by suppressing style semantics while preserving style fidelity.

\subsection{Semantic Leakage in Stylization}
Semantic leakage refers to the unintended transfer of style semantics to the generated image. 
In text-to-image stylization, recent methods suppress leakage through feature masking, negative guidance, patch suppression, or embedding purification~\cite{zhu2025less,Stylekeeper,Only-Style,CleanStyle}. 
In image-guided stylization, StyleSSP~\cite{Stylessp} decouples sampling from reference semantics, while StyleGallery~\cite{StyleGallery} constrains style matching via semantic correspondences. 
In contrast, SuRe directly regulates the native key--value correspondence to suppress semantic leakage, without additional model training.

\begin{figure*}[t]
    \centering
    \includegraphics[width= \textwidth]{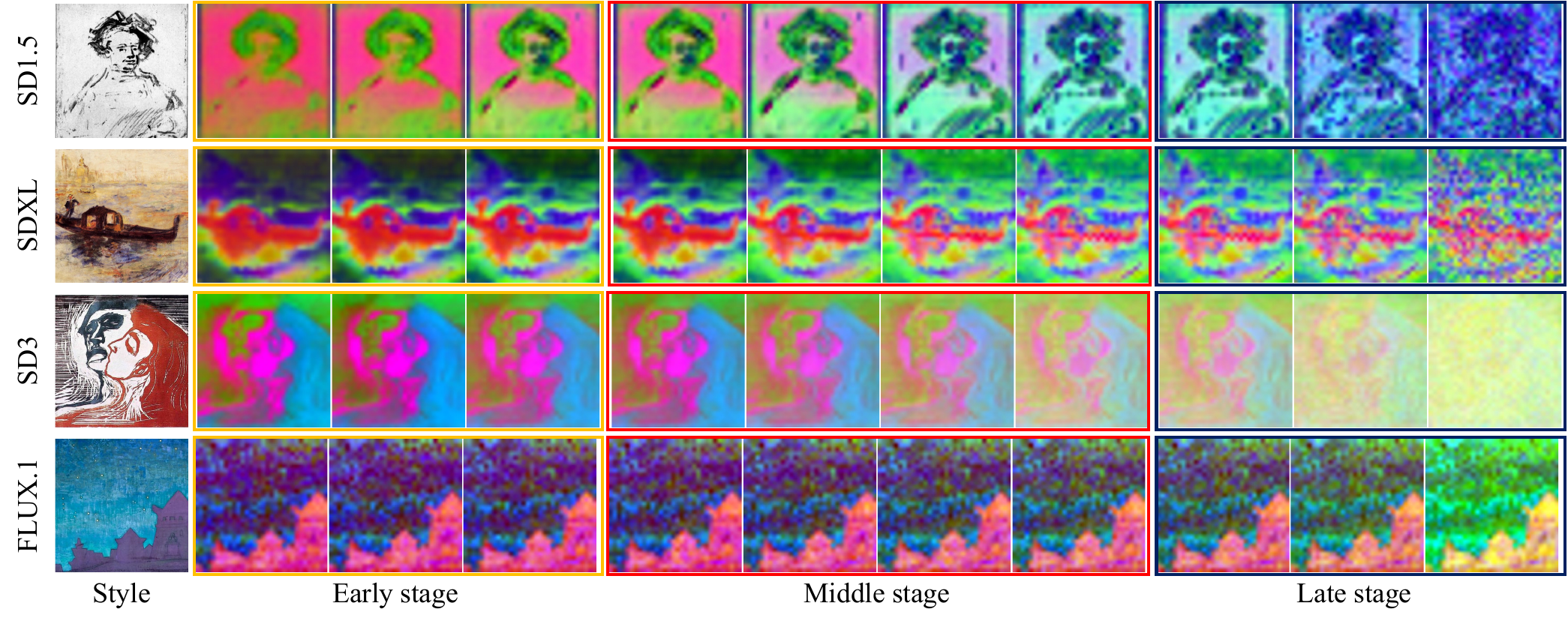}
    \caption{PCA visualization of $V_s^t$ extracted from the same $z_s^0$ at different timesteps. Coarse semantic layouts emerge in the early stage (first 30\%) and become clearer in the intermediate stage (middle 40\%), while high-frequency stylistic patterns gradually emerge in the late stage (last 30\%). We visualize $V_s^t$ from the 12th self-attention layer of SD1.5, the 45th of SDXL, the 19th of SD3, and the 51st of FLUX.}
    \label{fig:temporal_values}
\end{figure*}

\section{Method}
\label{sec:method}
In this section, we present SuRe for image-guided zero-shot stylization. 
We first propose V-Shuffle in Section~\ref{sec:vshuffle} to break the native key--value correspondence while preserving the empirical distribution of style values. 
Building on this, Section~\ref{sec:prm} introduces PRM for finer stylization. 
Finally, we extend SuRe to few-style inputs within the same framework in Section~\ref{sec:FWS}.
Fig.~\ref{fig:method} illustrates the overall architecture.
\subsection{Value Shuffle}
\label{sec:vshuffle}
Existing image-guided stylization methods typically inject style information by directly combining content queries with style keys and values at each sampling timestep $t$. 
However, the native spatial correspondence between $K_s^t$ and $V_s^t$ also preserves semantic associations inherited from the style image. 
Consequently, when the $i$-th content query matches the $j$-th style key, the corresponding value may introduce semantic information from the associated style region, leading to semantic leakage.

To break the key--value coupling while retaining the style information contained in $V_s^t$, we propose V-Shuffle, which randomly permutes the style values only along the spatial dimension. 
Specifically, a single shuffle operation is defined by applying a randomly sampled permutation matrix $P^t_l\in\mathbb{R}^{l \times l}$ to the $l$ dimension of $V_s^t\in\mathbb{R}^{h\times l\times d}$.
For the $a$-th attention head, it is formulated as
\begin{equation}
\tilde{V}_{s}^{t,a}=P^t_l \cdot V_{s}^{t,a},
\end{equation}
where $a \in \{1,\ldots,h\}$ and $V_s^{t,a}\in\mathbb{R}^{l\times d}$. 
Here, $h$ denotes the number of attention heads.
$d$ denotes the channel dimension. 
$l=h_r \times w_r$ represents the spatial dimension, i.e., the product of the height $h_r$ and width $w_r$ of the feature map at the corresponding resolution.
Notably, each element of $P^t_l$ is either 0 or 1, and every row and column of $P^t_l$ contains exactly one entry equal to 1, i.e., $P^t_l\mathbf{1}=\mathbf{1}$ and ${P_l^t}^\top\mathbf{1}=\mathbf{1}$.
For simplicity, we denote the head-wise application of V-Shuffle as $\tilde{V}_s^{t}=P^t_l \cdot V_s^t$. 
When multiple shuffles are used, we independently sample $m$ permutation matrices $\{P_l^{t,(i)}\}_{i=1}^{m}$, yielding
\begin{equation}
    \tilde{V}_s^{t,(i)} = P_l^{t,(i)} \cdot V_s^t, \quad i=1,\ldots,m.
\end{equation}

\begin{table*}[t]
\centering
\caption{Single-style transfer comparison. Top-4 results are highlighted in \textbf{\textcolor{red!80!black}{red}}, \textbf{\textcolor{blue!80!black}{blue}}, \textbf{\textcolor{orange!90!black}{orange}}, and \textbf{\textcolor{green!60!black}{green}}, respectively.}
\label{tab:style_transfer}
{\small
\setlength{\tabcolsep}{4.0pt}
\renewcommand{\arraystretch}{0.85}

\begin{tabular}{@{}>{\centering\arraybackslash}p{2.00cm} l c r r r r r r@{}}
\toprule

\multirow{2}{*}{Category} 
& \multirow{2}{*}{Method} 
& \multirow{2}{*}{Paradigm} 
& \multicolumn{3}{c}{Style Similarity} 
& \multicolumn{3}{c}{Content Similarity} \\

\cmidrule(lr){4-6} 
\cmidrule(l){7-9}

& & 
& ArtFID $\downarrow$ 
& FID $\downarrow$ 
& Gram $\downarrow$ 
& LPIPS $\downarrow$ 
& CFSD $\downarrow$ 
& DINOv2 $\uparrow$ \\

\midrule

\multirow{7}{*}{\shortstack[c]{Feed-forward\\methods}}
& AesPA-Net \venue{ICCV'23}{Aespa-net}
& CNN 
& 31.426 & 19.763 & 2.177 & 0.5135 & 0.2465 & 0.5119 \\

& EFDM \venue{CVPR'22}{EFDM}
& CNN 
& 30.146 & 17.344 & 3.061 & 0.6434 & 0.3134 & 0.4502 \\

& CAST \venue{SIGGRAPH'22}{CAST}
& CNN 
& 34.685 & 20.395 & 2.234 & 0.6212 & 0.2918 & 0.5587 \\

& StyTR$^2$ \venue{CVPR'22}{StyTR2}
& Transformer 
& 29.887 & 18.293 & 1.700 & 0.5491 & 0.2852 & 0.6098 \\

& AdaAttN \venue{ICCV'21}{AdaAttN}
& Transformer 
& 30.933 & 18.242 & 2.356 & 0.6076 & 0.3155 & 0.5036 \\

& MAST \venue{ACM MM'20}{MAST}
& CNN 
& 30.024 & 17.357 & 2.418 & 0.6356 & 0.2936 & 0.4255 \\

& AdaIN \venue{ICCV'17}{AdaIN}
& CNN 
& 29.705 & 17.527 & 2.161 & 0.6034 & 0.2946 & 0.4804 \\

\midrule

\multirow{5}{*}{\shortstack[c]{Diffusion\\adaptation\\methods}}
& AnyStyle \venue{ECCV'26}{AnyStyle}
& Diffusion 
& 32.360 
& \textbf{\textcolor{green!60!black}{16.497}}
& 2.283 
& 0.8494 
& 0.4626 
& 0.4842 \\

& OmniStyle \venue{CVPR'25}{omnistyle}
& Flow 
& 40.197 & 24.441 & 5.021 & 0.5800 & 0.3406 & 0.5930 \\

& K-LoRA \venue{CVPR'25}{K-lora}
& Diffusion 
& 47.979 & 28.088 & 3.027 & 0.6494 & 0.3826 & 0.4165 \\

& InstantStyle \venue{arXiv'24}{instantstyle}
& Diffusion 
& 40.409 & 23.225 & 2.845 & 0.6681 & 0.4608 & 0.4726 \\

& InST \venue{CVPR'23}{InST}
& Diffusion 
& 40.633 & 21.571 & 7.915 & 0.8002 & 0.4759 & 0.3716 \\

\midrule

\multirow{10}{*}{\shortstack[c]{Diffusion\\attention\\methods}}
& CoCoDiff \venue{ICLR'26}{cocodiff}
& Diffusion 
& 26.402 
& \textbf{\textcolor{orange!90!black}{15.226}} 
& \textbf{\textcolor{orange!90!black}{1.430}} 
& 0.6272 
& 0.3671 
& 0.3928 \\

& ACID-Style \venue{AAAI'26}{ACID-Style}
& Diffusion 
& 33.411 
& 20.946 
& 4.743 
& 0.5224 
& 0.2434 
& \textbf{\textcolor{blue!80!black}{0.7082}} \\

& StyleGallery \venue{CVPR'26}{StyleGallery}
& Diffusion 
& 27.614 
& 17.887 
& 2.804
& \textbf{\textcolor{green!60!black}{0.4639}}
& \textbf{\textcolor{green!60!black}{0.2261}}
& 0.6185 \\

& StyleSSP \venue{CVPR'25}{Stylessp}
& Diffusion 
& 34.932 
& 24.259 
& 4.226 
& 0.6205 
& 0.4200 
& 0.5382 \\

& AD \venue{CVPR'25}{AD}
& Diffusion 
& \textbf{\textcolor{green!60!black}{24.700}}
& \textbf{\textcolor{blue!80!black}{14.763}} 
& 1.903 
& 0.5670 
& 0.2755 
& 0.4987 \\

& StyleID \venue{CVPR'24}{StyleID}
& Diffusion 
& 28.801 
& 18.131 
& 1.732 
& 0.5055 
& 0.2281 
& 0.5628 \\

\cmidrule(l){2-9}

& \cellcolor{blue!8}\textbf{Ours (SD1.5)}
& \cellcolor{blue!8}\textbf{Diffusion}
& \cellcolor{blue!8}\textbf{\textcolor{red!80!black}{21.714}}
& \cellcolor{blue!8}\textbf{\textcolor{red!80!black}{13.833}}
& \cellcolor{blue!8}1.621
& \cellcolor{blue!8}\textbf{\textcolor{orange!90!black}{0.4620}}
& \cellcolor{blue!8}\textbf{\textcolor{blue!80!black}{0.2049}}
& \cellcolor{blue!8}\textbf{\textcolor{green!60!black}{0.6459}} \\

& \cellcolor{blue!8}\textbf{Ours (SDXL)}
& \cellcolor{blue!8}\textbf{Diffusion}
& \cellcolor{blue!8}26.120
& \cellcolor{blue!8}20.403
& \cellcolor{blue!8}\textbf{\textcolor{green!60!black}{1.482}}
& \cellcolor{blue!8}0.5941
& \cellcolor{blue!8}0.2524
& \cellcolor{blue!8}0.5115 \\

& \cellcolor{blue!8}\textbf{Ours (SD3)}
& \cellcolor{blue!8}\textbf{Flow}
& \cellcolor{blue!8}\textbf{\textcolor{blue!80!black}{23.275}}
& \cellcolor{blue!8}17.417
& \cellcolor{blue!8}\textbf{\textcolor{blue!80!black}{1.260}}
& \cellcolor{blue!8}\textbf{\textcolor{red!80!black}{0.3724}}
& \cellcolor{blue!8}\textbf{\textcolor{red!80!black}{0.1559}}
& \cellcolor{blue!8}\textbf{\textcolor{red!80!black}{0.7169}} \\

& \cellcolor{blue!8}\textbf{Ours (FLUX)}
& \cellcolor{blue!8}\textbf{Flow}
& \cellcolor{blue!8}\textbf{\textcolor{orange!90!black}{24.272}}
& \cellcolor{blue!8}17.169
& \cellcolor{blue!8}\textbf{\textcolor{red!80!black}{0.788}}
& \cellcolor{blue!8}\textbf{\textcolor{blue!80!black}{0.4460}}
& \cellcolor{blue!8}\textbf{\textcolor{orange!90!black}{0.2202}}
& \cellcolor{blue!8}\textbf{\textcolor{orange!90!black}{0.6896}} \\

\bottomrule
\end{tabular}
}
\end{table*}

\begin{figure*}[t]
    \centering
    \includegraphics[width=\textwidth]{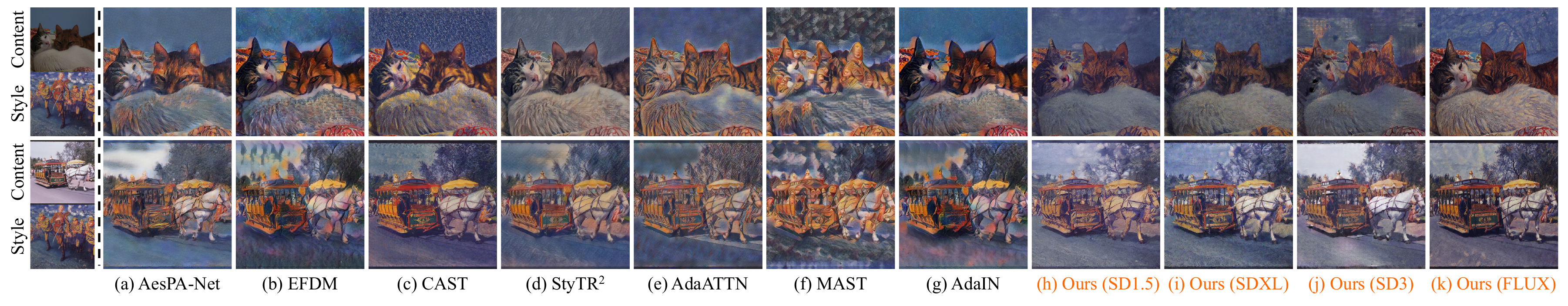}
    \caption{Comparison with feed-forward methods. SuRe preserves content structure while transferring reference colors and textures.}
    \label{fig:comparison_feedforward}
\end{figure*}

\paragraph{Empirical Distribution Preservation.} An important property of V-Shuffle is that it exactly preserves the empirical distribution of the value tokens and breaks the native key--value correspondence without requiring any additional training.
For the $a$-th attention head, we regard $V_s^{t,a}$ as a collection of $l$ $d$-dimensional value tokens, where $V_s^{t,a,j}\in\mathbb{R}^{d}$ denotes the $j$-th token.
The empirical distribution of these value tokens is then defined as
\begin{equation}
\widehat{\mu}_{V_s^{t,a}}
=
\frac{1}{l}\sum_{j=1}^{l}
\delta_{V_s^{t,a,j}},
\label{eq:value_distribution}
\end{equation}
where $\delta_{V_s^{t,a,j}}$ denotes the Dirac measure centered at $V_s^{t,a,j}$.
Accordingly, the empirical distribution of the shuffled values $\tilde{V}_s^{t,a}$ is
\begin{equation}
\widehat{\mu}_{\tilde{V}_s^{t,a}}
=
\frac{1}{l}\sum_{k=1}^{l}
\delta_{\tilde{V}_s^{t,a,k}}.
\end{equation}

Let $\pi$ denote the permutation induced by $P^t_l$ and $k=\pi(j)$ denote the new position of the original $j$-th token after permutation. 
Then, $\tilde{V}_s^{t,a,\pi(j)} = V_s^{t,a,j}$.
Since $\pi$ is a bijection over $\{1,\ldots,l\}$, the shuffled positions
$\{\pi(j)\}_{j=1}^{l}$ still cover all spatial positions exactly once.
Therefore,
\begin{equation}
\widehat{\mu}_{\tilde{V}_s^{t,a}}
=
\frac{1}{l}\sum_{j=1}^{l}
\delta_{\tilde{V}_s^{t,a,\pi(j)}}
=
\frac{1}{l}\sum_{j=1}^{l}
\delta_{V_s^{t,a,j}}
=
\widehat{\mu}_{V_s^{t,a}}.
\label{eq:distribution_invariance}
\end{equation}
This empirical distribution-preserving property enables V-Shuffle to break the native key--value correspondence without optimizing the underlying value features.
Numerical verification of this preservation property and ablations of its importance are provided in \textbf{Supp.}~\ref{sec:value_distribution_verification}.
Notably, preserving the empirical distribution of the value tokens does not imply preserving the distribution of the resulting attention, because V-Shuffle changes the correspondence between the attention weights and value tokens, thereby altering the weighted aggregation. 
A proof is provided in \textbf{Supp.}~\ref{sec:attention_output_change}.

\begin{figure*}[t]
    \centering
    \includegraphics[width=\textwidth]{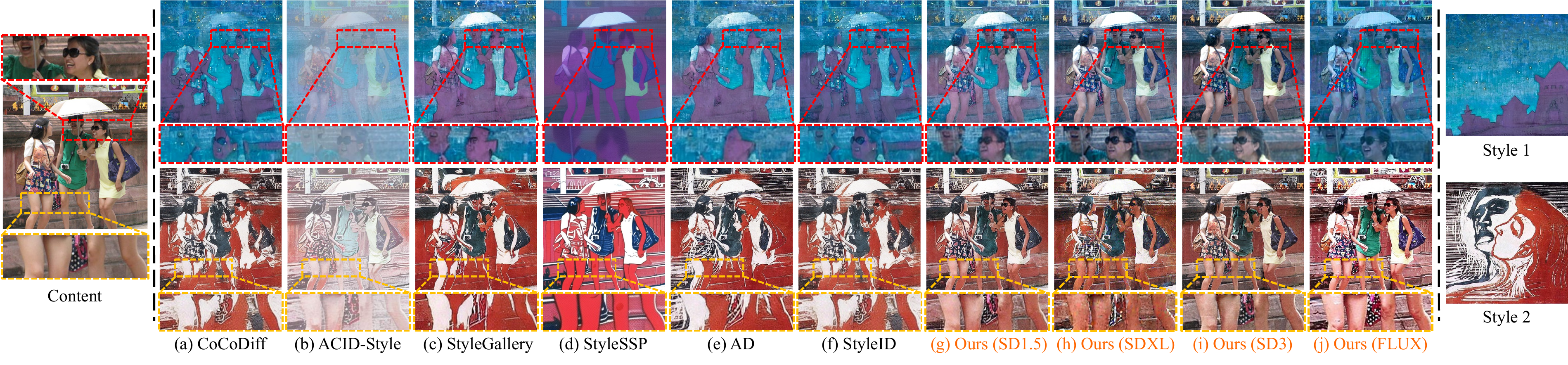}
    \caption{Comparison with diffusion attention methods. SuRe suppresses semantic leakage while preserving style.}
    \label{fig:comparison_attention}
\end{figure*}

\begin{figure}[t]
    \centering
    \includegraphics[width=0.8\linewidth]{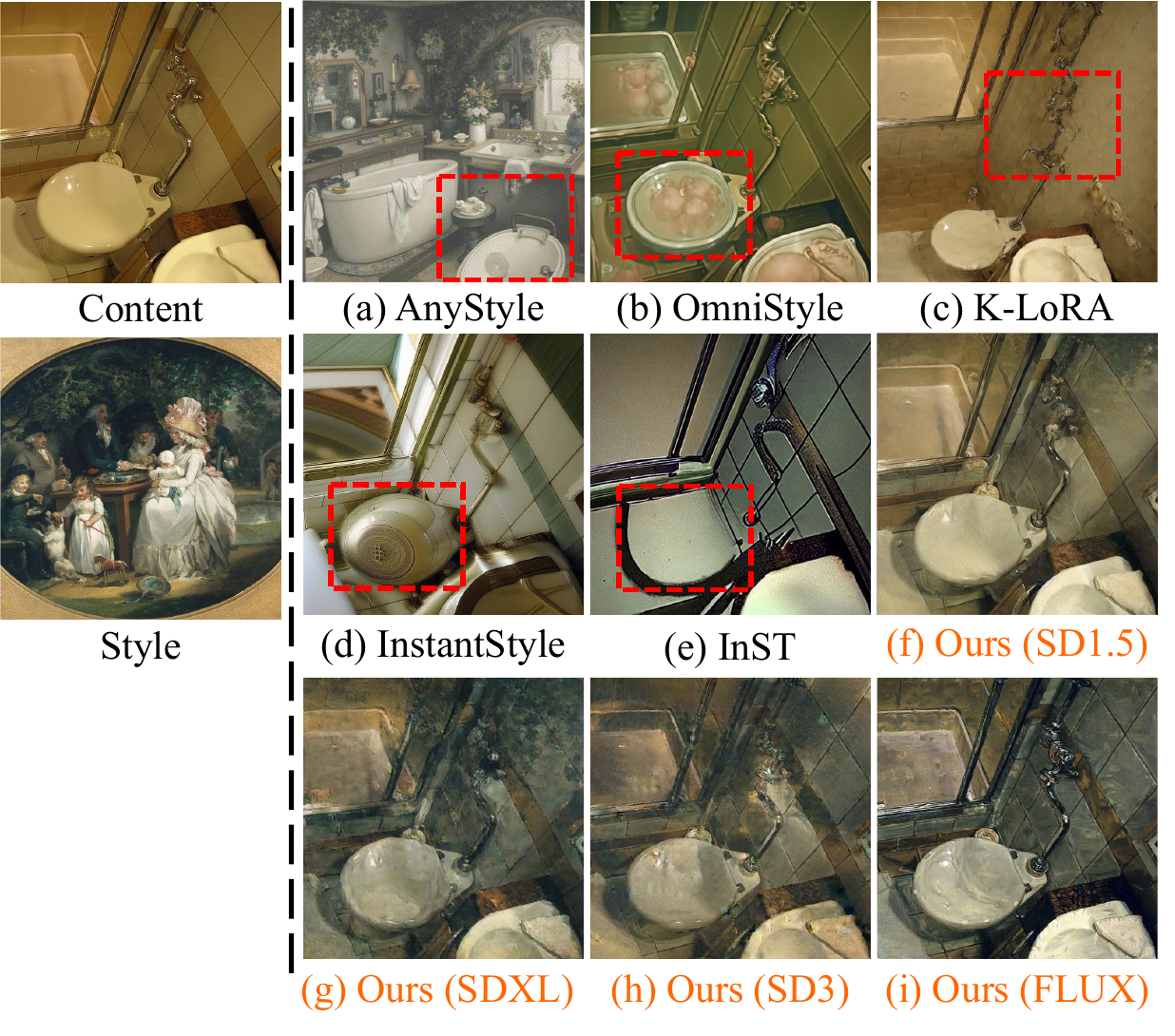}
    \caption{Comparison with diffusion adaptation methods. SuRe preserves the bathroom layout and transfers the painterly style.}
    \label{fig:comparison_adaptation}
\end{figure}

\subsection{Progressive Recoupling Mechanism}
\label{sec:prm}
Dynamic denoising exhibits time-dependent behaviors, raising the question of whether style values follow a similar pattern during stylization. 
Unlike~\cite{cocodiff,StyleGallery,StyleID}, which rely on DDIM inversion~\cite{DDIM} to obtain latent trajectories, we keep the style latent $z_s^0$ and content latent $z_c^0$ fixed throughout stylization and initialize the stylized latent from $z_c^0$ (further analysis of this design choice is provided in \textbf{Supp.}~\ref{sec:fixed_latent_analysis}). 
We therefore analyze the style values $V_s^t$ extracted from $z_s^0$ at different timesteps using principal component analysis (PCA). Interestingly, as shown in Fig.~\ref{fig:temporal_values}, $V_s^t$ still exhibits clear timestep-dependent patterns without adding noise to $z_s^0$: semantics are more prominent at earlier timesteps, while high-frequency stylistic patterns become increasingly evident at later timesteps. This suggests that stronger V-Shuffle should be applied at early timesteps to suppress semantic transfer, followed by progressive restoration of the native correspondence for finer stylization.

To achieve this, we introduce PRM and control the shuffling strength using a cosine weight:
\begin{equation}
\alpha_t=\frac{1}{2}\left[1+\cos\left(\frac{\pi t}{T} \right)\right],
\label{eq:cosine_schedule}
\end{equation}
where $t$ denotes the current timestep and $T$ is the total number of sampling timesteps. To implement progressive recoupling, we optimize the stylized latent via attention distillation~\cite{AD}.
Let $z^t_{cs}$ denote the stylized latent at timestep $t$, with $Q_{cs}^t$, $K_{cs}^t$, and $V_{cs}^t$ denoting its corresponding query, key, and value. 
We guide its attention output toward both the shuffled and native style attention, and progressively balance their contributions using the cosine weight $\alpha_t$. 
The style recoupling loss is defined as
\begin{equation}
\begin{aligned}
    \mathcal{L}_{s} &= \alpha_t \underbrace{||\textbf{A}(Q_{cs}^t,K_{cs}^t,V_{cs}^t) - \textbf{A}(Q_c^t, K_s^t, V_s^t)||_1}_{\text{Unshuffled term}} \\
    &+ (1-\alpha_t) \underbrace{\frac{1}{m} \sum_{i=1}^{m} \left\| \textbf{A}(Q_{cs}^t,K_{cs}^t,V_{cs}^t)  - \textbf{A}(Q_c^t, K_s^t, \tilde{V}_s^{t,(i)}) \right\|_1}_{\text{Shuffled term}}.
    \label{PRM}
\end{aligned}
\end{equation}
Here, $\textbf{A}$ denotes scaled dot-product attention, and $m$ is the number of permutation matrices at each timestep.
Averaging over the independently sampled permutations yields an unbiased Monte Carlo estimate of the expected attention-matching loss under random value shuffling.
This encourages stylization across different value assignments rather than reliance on a single shuffled attention target.

As sampling proceeds from $t=T$ to $t=0$, $\alpha_t$ increases monotonically from $0$ to $1$. Therefore, the shuffled term remains dominant in the early stage, emphasizing style semantic uncoupling.
During the intermediate stage, $\alpha_t$ increases more rapidly, progressively shifting the guidance toward the unshuffled term.
In the late stage, $\alpha_t$ approaches $1$ smoothly, allowing native key--value correspondence to guide high-frequency stylistic textures.
This yields a gradual transition from semantic uncoupling to style recoupling throughout sampling. 
Finally, following~\cite{AD}, we apply the content loss $\mathcal{L}_{c}=||Q_{cs}^t - Q_c^t||_1$, and formulate the overall objective as
\begin{equation}
    \mathcal{L}_{total} = \mathcal{L}_{s} + \beta \cdot \mathcal{L}_{c}.
\end{equation}
Algorithm pseudocode is provided in \textbf{Supp.}~\ref{sec:algorithm_supp}.

\begin{table}[t]
    \centering
    \caption{Component ablation. Best results are highlighted in \textbf{\textcolor{red!80!black}{red}}.}
    \label{tab:ablation1}
    {\footnotesize
    \setlength{\tabcolsep}{2.0pt}
    \renewcommand{\arraystretch}{0.80}

    \begin{tabular}{lcccccc}
        \toprule
        \multirow{2}{*}{Configuration}
        & \multicolumn{3}{c}{Style Similarity}
        & \multicolumn{3}{c}{Content Similarity} \\
        \cmidrule(lr){2-4} \cmidrule(lr){5-7}
        & ArtFID $\downarrow$
        & FID $\downarrow$
        & Gram $\downarrow$
        & LPIPS $\downarrow$
        & CFSD $\downarrow$
        & DINOv2 $\uparrow$ \\
        \midrule

        Baseline
        & 24.700
        & 14.763
        & 1.903
        & 0.5670
        & 0.2755
        & 0.4987 \\
        \midrule
        
        + V-Shuffle
        & 30.815
        & 20.296
        & 3.567
        & \textbf{\textcolor{red!80!black}{0.4470}}
        & \textbf{\textcolor{red!80!black}{0.1996}}
        & \textbf{\textcolor{red!80!black}{0.6787}} \\

        + PRM
        & \textbf{\textcolor{red!80!black}{21.714}}
        & \textbf{\textcolor{red!80!black}{13.833}}
        & \textbf{\textcolor{red!80!black}{1.621}}
        & 0.4620
        & 0.2049
        & 0.6459 \\

        \bottomrule
    \end{tabular}
    }
\end{table}

\begin{table}[t]
    \centering
    \caption{Effect of the number of permutation matrices $m$. Top-3 results are highlighted in \textbf{\textcolor{red}{red}}, \textbf{\textcolor{blue}{blue}}, and \textbf{\textcolor{orange!90!black}{orange}}, respectively.}
    \label{tab:permutation_number}
    \footnotesize
    \setlength{\tabcolsep}{2.0pt}
    \renewcommand{\arraystretch}{0.80}
    \begin{tabular}{cccccccc}
        \toprule
        $m$ & ArtFID $\downarrow$ & FID $\downarrow$ & Gram $\downarrow$ & LPIPS $\downarrow$ & CFSD $\downarrow$ & DINOv2 $\uparrow$ & Time $\downarrow$  \\
        \midrule
        1 & 24.644 & 15.797 & 1.735 & 0.4672 & 0.2146 & \textbf{\textcolor{blue}{0.6468}} & \textbf{\textcolor{red}{42.26s}}\\
        3 & 22.415 & 14.301 & 1.689 & 0.4650 & 0.2146 & \textbf{\textcolor{orange!90!black}{0.6463}} & \textbf{\textcolor{blue}{45.91s}} \\
        5 & \textbf{\textcolor{orange!90!black}{21.714}}
          & \textbf{\textcolor{blue}{13.833}}
          & \textbf{\textcolor{blue}{1.621}}
          & \textbf{\textcolor{orange!90!black}{0.4620}}
          & \textbf{\textcolor{orange!90!black}{0.2049}}
          & 0.6459 & \textbf{\textcolor{orange}{47.69s}} \\
        7 & \textbf{\textcolor{blue}{21.544}}
          & \textbf{\textcolor{red}{13.832}}
          & \textbf{\textcolor{red}{1.616}}
          & \textbf{\textcolor{blue}{0.4605}}
          & \textbf{\textcolor{red}{0.2044}}
          & 0.6458
          & 57.15s \\
        9 & \textbf{\textcolor{red}{21.543}}
          & \textbf{\textcolor{orange!90!black}{13.834}}
          & \textbf{\textcolor{orange!90!black}{1.626}}
          & \textbf{\textcolor{red}{0.4525}}
          & \textbf{\textcolor{blue}{0.2045}}
          & \textbf{\textcolor{red}{0.6471}}
          & 60.32s \\
        \bottomrule
    \end{tabular}
\end{table}

\subsection{Few-Style Shuffling}
\label{sec:FWS}
Our framework naturally extends to a few style images from the same style domain.
Specifically, given $n$ style images $I_{s_{1:n}}$, we extract their keys and values and concatenate them along the token dimension to obtain $K_{s_{1:n}}^t$ and $V_{s_{1:n}}^t$.
We then apply V-Shuffle along the spatial dimension of the concatenated value features, yielding $\tilde{V}_{s_{1:n}}^t$, while keeping $K_{s_{1:n}}^t$ unchanged.
The resulting keys and values replace their single-style counterparts in Eq.~\ref{PRM}, with $\tilde{V}_{s_{1:n}}^t$ used in the shuffled term and $V_{s_{1:n}}^t$ in the unshuffled term.
PRM follows the same cosine schedule to progressively regulate the contributions of these two terms throughout sampling.

\begin{figure}[t]
    \centering
    \includegraphics[width=0.90\linewidth]{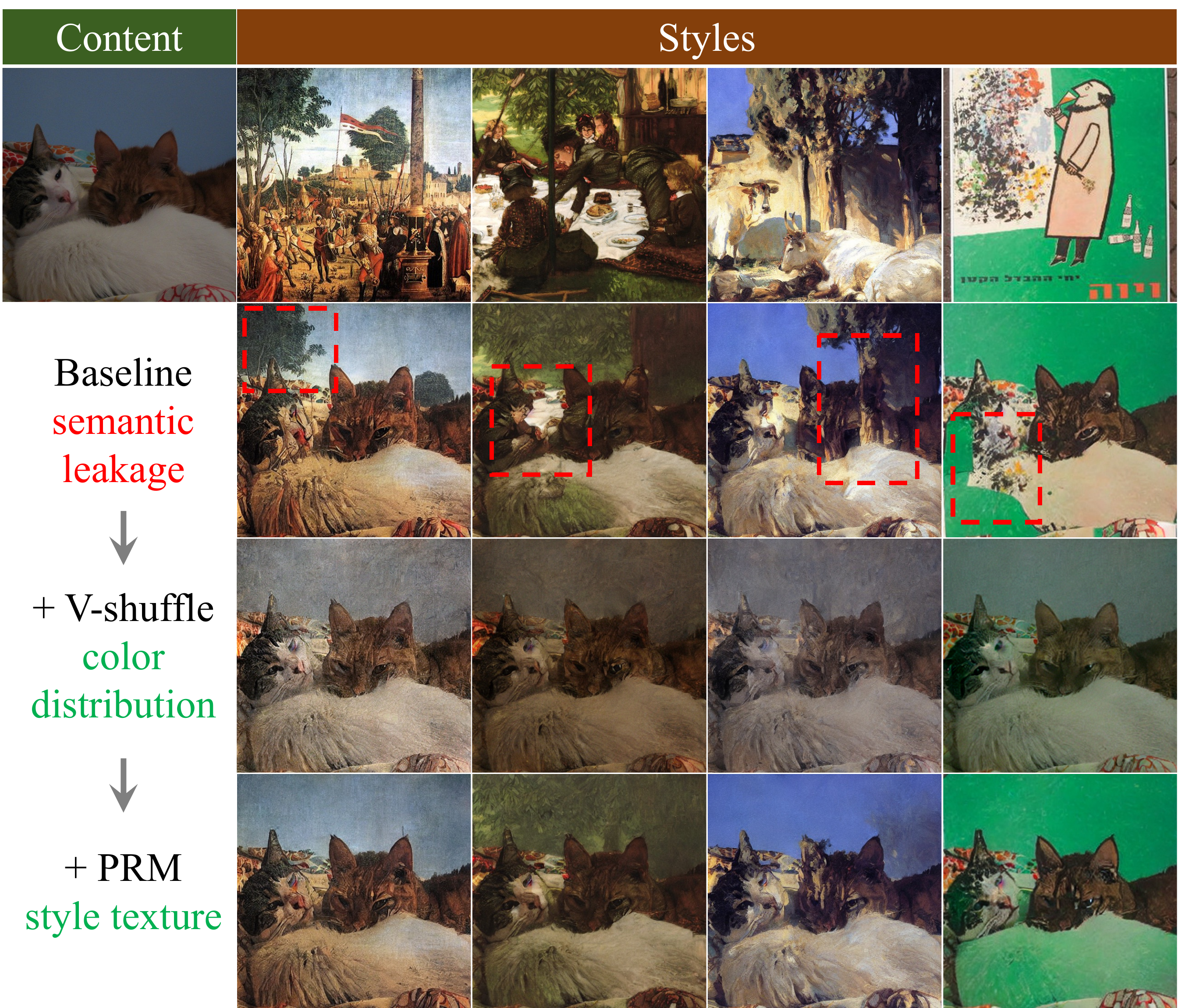}
    \caption{Qualitative analysis of each component. V-Shuffle suppresses semantic leakage while preserving colors, and PRM restores stylistic textures. \textcolor{red}{Red boxes} indicate semantic leakage.}
    \label{fig:ablation1}
\end{figure}

\section{Experiments}
\label{sec:method}
\subsection{Experiment Setup}
\textbf{Evaluation Datasets and Metrics.} We evaluate single-style transfer on the benchmark used in~\cite{StyleID,ACID-Style,StyTR2}, containing 20 MS-COCO~\cite{lin2014microsoft} content images and 40 WikiArt~\cite{tan2018improved} style images, yielding 800 stylized images.
Following~\cite{StyleID,cocodiff}, we use LPIPS and CFSD for content similarity, FID and ArtFID for style similarity.
Additionally, we use Gram loss to assess style consistency and DINOv2~\cite{dinov2} to measure semantic consistency.

\textbf{Implementation Details.}
We use Stable Diffusion (SD) 1.5~\cite{LDM} as the default backbone and further evaluate SuRe on SDXL~\cite{SDXL}, SD3~\cite{SD3}, and FLUX.1-dev~\cite{FLUX}.
For simplicity, FLUX.1-dev is denoted as FLUX in all figures and tables.
For SD1.5, we use the last six attention layers and perform $T=200$ optimization steps on the stylized latent at $512\times512$ resolution.
At each step, we optimize $z^t_{cs}$ using Adam with a learning rate of $0.05$ and bfloat16 precision on a single NVIDIA A100 GPU, with $\beta=0.24$ and $m=5$.
Hyperparameters for the other backbones are provided in \textbf{Supp.}~\ref{sec:backbone_hyperparameters}.
All ablation experiments use SD1.5.

\subsection{Comparison with the State-of-the-Art}
We compare SuRe with 18 representative image-guided methods, including conventional feed-forward methods (StyTR$^2$~\cite{StyTR2}, AdaAttN~\cite{AdaAttN}, AesPA-Net~\cite{Aespa-net}, CAST~\cite{CAST}, EFDM~\cite{EFDM}, MAST~\cite{MAST}, and AdaIN~\cite{AdaIN}), diffusion adaptation methods (AnyStyle~\cite{AnyStyle}, InstantStyle~\cite{instantstyle}, OmniStyle~\cite{omnistyle}, InST~\cite{InST}, and K-LoRA~\cite{K-lora}), and diffusion attention methods (CoCoDiff~\cite{cocodiff}, ACID-Style~\cite{ACID-Style}, StyleGallery~\cite{StyleGallery}, StyleSSP~\cite{Stylessp}, AD~\cite{AD}, and StyleID~\cite{StyleID}).
We further provide an efficiency comparison and a user study in \textbf{Supp.}~\ref{sec: efficiency comparison} and \textbf{Supp.}~\ref{sec:user_study}, respectively.

\textbf{Quantitative Analysis.}
Tab.~\ref{tab:style_transfer} shows that our method performs strongly across all six metrics, with most results ranking in the top three across four backbones. Specifically, Ours (SD1.5) achieves the best ArtFID and FID, while Ours (FLUX) obtains the best Gram score. The improved content similarity further indicates that SuRe alleviates semantic leakage without sacrificing style fidelity.

\textbf{Qualitative Analysis.}
Fig.~\ref{fig:comparison_feedforward} compares SuRe with feed-forward methods. 
Existing methods often introduce inconsistent styles or visual artifacts, while SuRe better preserves content structures and transfers the reference style across all backbones. 
In Fig.~\ref{fig:comparison_adaptation}, adaptation methods may introduce visual hallucinations, such as altered layouts and deformed objects, whereas SuRe maintains the original structure while reproducing the reference colors and textures. Fig.~\ref{fig:comparison_attention} compares SuRe with existing diffusion attention methods, which still suffer from semantic leakage to varying degrees, leading to missing or distorted facial and leg structures. SuRe better preserves these details while maintaining the corresponding painterly style.

\begin{figure}[t]
    \centering
    \includegraphics[width=0.90\linewidth]{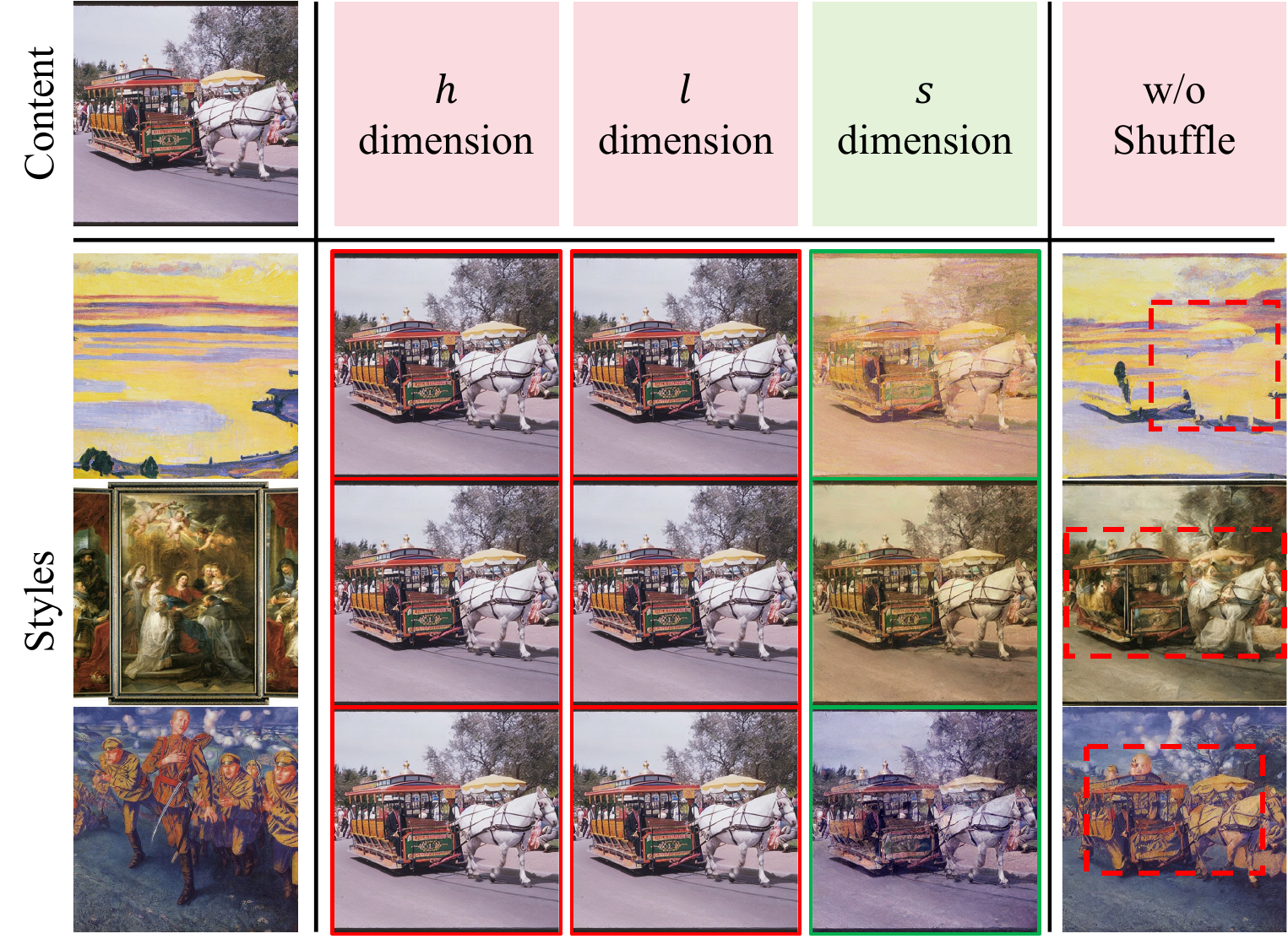}
    \caption{Ablation on the shuffle dimension. Shuffling along the $l$ dimension suppresses semantic leakage while preserving style cues. \textcolor{red}{Red dashed boxes} indicate leaked semantics.}
    \label{fig:toy}
\end{figure}

\subsection{Ablation Studies}
\subsubsection{Component Ablation}
Tab.~\ref{tab:ablation1} and Fig.~\ref{fig:ablation1} demonstrate the complementary effects of V-Shuffle and PRM. 
V-Shuffle suppresses semantic leakage while preserving low-level color distributions, improving all three content similarity metrics.
PRM further incorporates fine-grained stylistic textures, improving FID, Gram loss, and ArtFID, while retaining better content similarity than the baseline across all three metrics. 
These results demonstrate a progression from semantic uncoupling to fine-grained style recoupling.

\subsubsection{Number of Permutation Matrices}
As shown in Tab.~\ref{tab:permutation_number}, increasing $m$ provides a more stable Monte Carlo estimate and improves stylization quality. 
However, the performance saturates when $m \geq 5$, while the inference time continues to increase. 
Thus, we set $m=5$ as a good trade-off between performance and efficiency.

\subsubsection{Selection of Shuffle Dimension}
We ablate the shuffle dimension by applying permutation matrices to the $h$, $l$, and $d$ dimensions of $V_s^t$, respectively, without PRM. 
As shown in Fig.~\ref{fig:toy}, shuffling along $h$ or $d$ removes most style information, producing outputs close to the content image. 
In contrast, shuffling along the spatial dimension $l$ effectively suppresses semantic leakage while preserving low-level color distributions. 
Therefore, we adopt $l$ as the default shuffle dimension.

\begin{figure}[t]
    \centering
    \includegraphics[width=\linewidth]{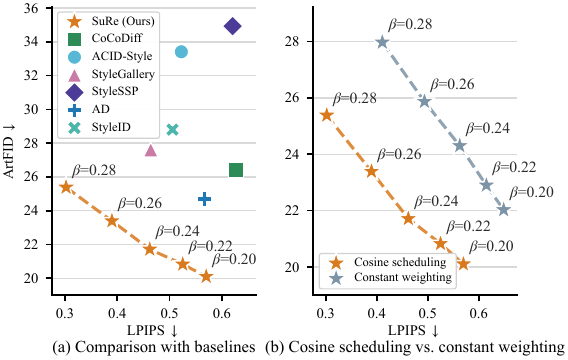}
    \caption{Ablation on different $\beta$ and injection strategies.}
    \label{fig:ablation3}
\end{figure}

\subsubsection{Study of $\beta$}
$\beta$ controls the trade-off between the style loss $\mathcal{L}_{s}$ and the content loss $\mathcal{L}_c$. To demonstrate the effectiveness of our method, we plot the Pareto curves under different $\beta$ values. As shown in Fig.~\ref{fig:ablation3}(a), compared with baselines, our curve consistently lies in the lower-left region relative to other methods, achieving a superior Pareto frontier. This result further validates the effectiveness of our method.

\subsubsection{Different Injection Strategies}
Our method optimizes the stylized latent at each step using a style loss $\mathcal{L}_{s}$ that combines shuffled and unshuffled terms with cosine scheduling. To validate this design, we compare two alternatives: Constant weighting (COW) and direct replacement (DR). COW linearly weights the two terms with a fixed $\alpha_t=0.5$, while DR directly replaces $\textbf{A}(Q_{cs}^t,K_{cs}^t,V_{cs}^t)$ with $\textbf{A}(Q_c^t,K_s^t,\tilde{V}_s^t)$ without loss-based optimization.

\begin{figure}[t]
    \centering
    \includegraphics[width=0.95\linewidth]{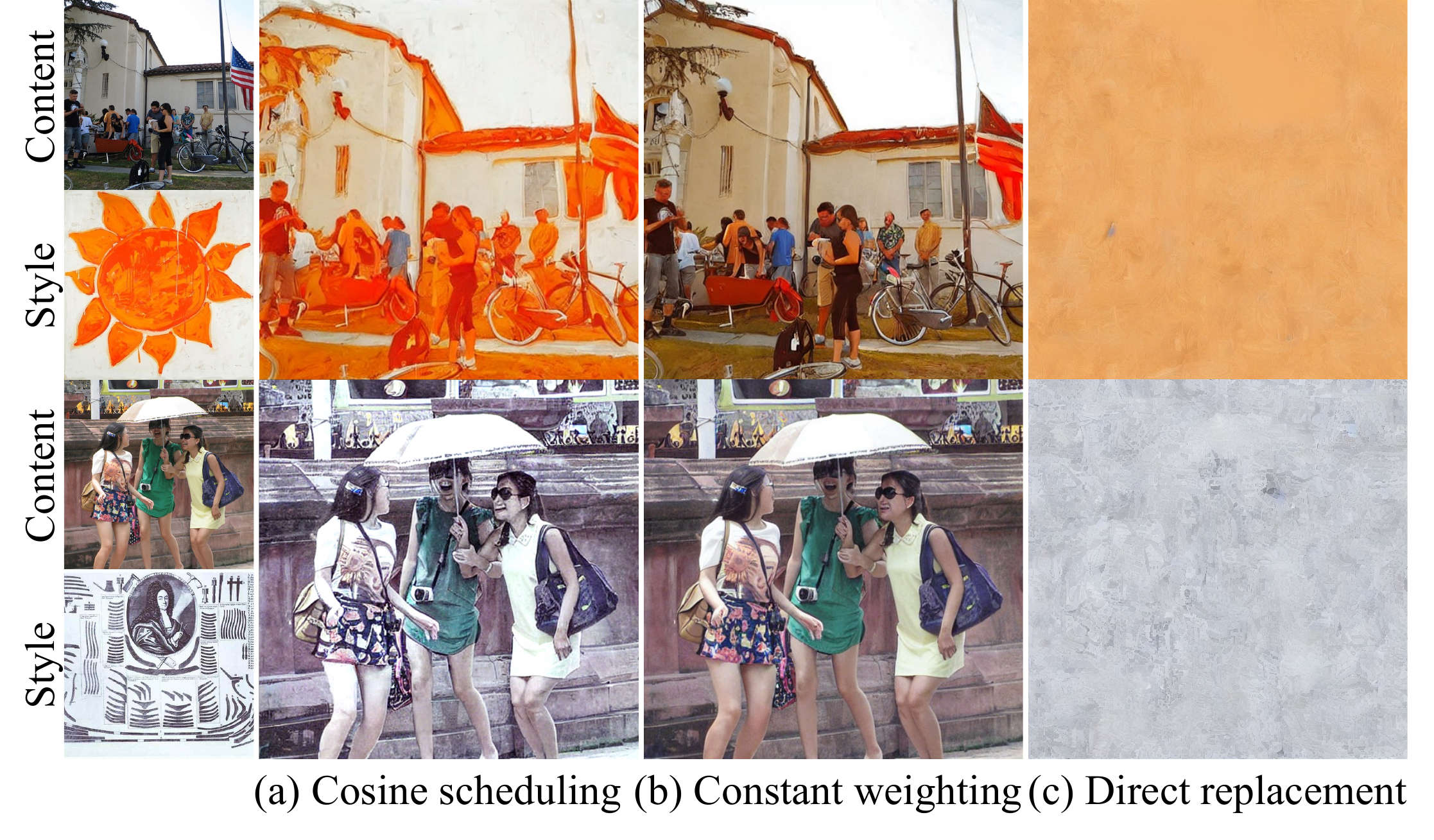}
    \caption{Visual comparison of different injection strategies.}
    \label{fig:ablation4}
\end{figure}

\begin{figure}[h]
    \centering
    \includegraphics[width=\linewidth]{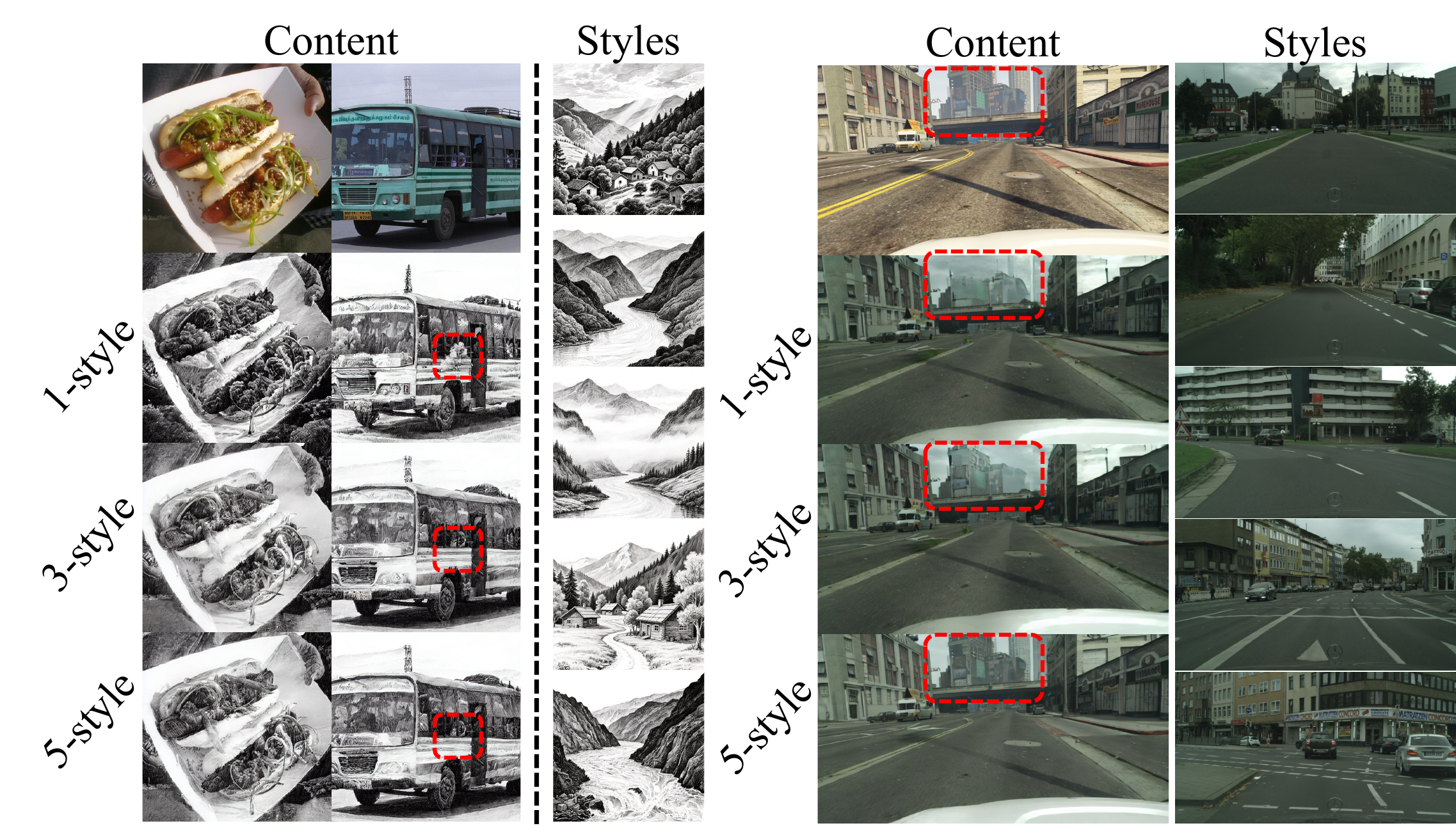}
    \caption{Qualitative comparison with different numbers of style images. \textcolor{red}{Red dashed boxes} highlight reduced semantic leakage.}
    \label{fig:multi}
\end{figure}

As shown in Fig.~\ref{fig:ablation4}, COW yields weaker stylization due to the lack of time-aware scheduling. Fig.~\ref{fig:ablation3}(b) further shows that our Pareto curve consistently lies in the lower-left region relative to COW under different $\beta$ values. DR mainly transfers color distributions because shuffling disrupts the semantics in $\tilde{V}_s^t$, resulting in limited structural preservation. In contrast, our loss-based optimization constrains semantic structure with the content loss while progressively regulating style through PRM, enabling more natural and visually pleasing stylization.

\subsection{Multiple Style Images}
We further observe that using multiple style images from the same domain can suppress semantic leakage and encourage domain-invariant style learning. To evaluate this setting, we use multiple images from the same style domain as references and qualitatively compare different numbers of style images. Since existing style-similarity metrics are designed for a single reference, we conduct only qualitative analysis. As shown in Fig.~\ref{fig:multi}, increasing the number of style images from one to five improves content preservation without sacrificing style fidelity. This suggests that shuffling values across multiple style images encourages the model to capture shared style statistics rather than image-specific patterns, yielding more domain-invariant style representations and reduced semantic leakage.

\section{Conclusion}
We presented SuRe for mitigating semantic leakage in image-guided zero-shot stylization. 
Our key observation is that the native correspondence between style keys and values couples stylistic and semantic information during attention aggregation. 
We thus propose V-Shuffle to break this correspondence by permuting style values while preserving their empirical distribution, thereby suppressing unintended semantic transfer. 
PRM then gradually restores the native correspondence during latent stylization optimization to recover fine‑grained stylistic textures.
Experiments show that SuRe improves the balance between content preservation and style fidelity across multiple diffusion backbones, and naturally extends to multi-style transfer.
\section*{LLM Usage Statement}
In this work, OpenAI Codex (GPT-6 Astra) is used to assist with code development, implementation refinement, and debugging, including identifying and resolving programming errors and improving code organization. 
It also supports manuscript preparation through English language polishing, grammatical corrections, improving clarity and readability, and LaTeX formatting. 
The authors review and verify all AI-assisted content and take full responsibility for the research and its conclusions.
{
    \small
    \bibliographystyle{ieeenat_fullname}
    \bibliography{main}
}

\clearpage
\setcounter{page}{1}
\maketitlesupplementary

\section{Linear Probing Implementation Details}
\label{sec:rationale}
In the introduction, we conduct a linear probing experiment on SD1.5 to investigate whether semantic leakage originates from key--value coupling.
Since no suitable dataset in image-guided stylization directly measures the degree of semantic leakage, we use two datasets commonly adopted in sim-to-real image translation, GTA5~\cite{GTA5} and Cityscapes~\cite{cordts2016cityscapesdatasetsemanticurban}.
Specifically, we sample 10 images from each dataset, using GTA5 as content images and Cityscapes as style images.
The first seven images from each set form $7\times7$ training pairs, while the remaining three form $3\times3$ test pairs.
For each of the three settings introduced in the introduction, we extract self-attention output features over 200 steps and six layers, yielding $49\times200\times6=58{,}800$ training samples and $9\times200\times6=10{,}800$ test samples.
A separate linear probe is trained and evaluated for each setting.
The diffusion backbone and all extracted attention features remain frozen, so that the probe measures only semantic information already encoded in the attention output.

\paragraph{Why does joint shuffling produce results nearly identical to native attention?} As shown in Fig.~\ref{fig:LR}, Native $K$--$V$ pairing and Joint $K$--$V$ reordering produce nearly identical results. Here, we provide a mathematical analysis to explain this observation. Let scaled dot-product attention be
\begin{equation}
\mathbf{A}(Q_c,K_s,V_s)=\operatorname{Softmax}\!\left(\frac{Q_cK_s^{\top}}{\sqrt{d}}\right)V_s,
\label{eq:attention_supp}
\end{equation}
where the softmax is applied row-wise. Let $P_l$ be a permutation matrix. Since right multiplication by $P_l^{\top}$ only reorders the entries of every row, row-wise softmax is permutation equivariant:
\begin{equation}
\operatorname{Softmax}(XP_l^{\top})
= \operatorname{Softmax}(X)P_l^{\top}.
\label{eq:softmax_equivariance}
\end{equation}
Therefore, jointly reordering keys and values leaves attention unchanged:
\begin{align}
\mathbf{A}(Q_c,P_lK_s,P_lV_s)
&=\operatorname{Softmax}\!\left(\frac{Q_cK_s^{\top}P_l^{\top}}{\sqrt{d}}\right)P_lV_s \\
&=\operatorname{Softmax}\!\left(\frac{Q_cK_s^{\top}}{\sqrt{d}}\right)P_l^{\top}P_lV_s \\
&=\operatorname{Softmax}\!\left(\frac{Q_cK_s^{\top}}{\sqrt{d}}\right)V_s \\
&=\mathbf{A}(Q_c,K_s,V_s),
\label{eq:joint_shuffle_invariance}
\end{align}
where $P_l^{\top}P_l=I$. This explains why the Native $K$--$V$ pairing and Joint $K$--$V$ reordering yield nearly identical probing performance in Fig.~\ref{fig:LR}.

\begin{table}[t]
    \centering
    \caption{Numerical verification of empirical value-distribution preservation under V-Shuffle, using one representative content--style pair across 200 sampling steps and six selected attention layers (1,200 layer--timestep pairs). ``Max. error'' is the maximum over all evaluated timesteps, layers, attention heads, and channels.}
    \label{tab:value_distribution_verification}
    \small
    \setlength{\tabcolsep}{2.5pt}
    \renewcommand{\arraystretch}{0.80}

    \begin{tabular}{
        >{\centering\arraybackslash}m{1.55cm}
        >{\centering\arraybackslash}m{4.05cm}
        >{\centering\arraybackslash}m{1.70cm}
    }
        \toprule
        Statistic & Definition & Max. error \\
        \midrule

        \shortstack[c]{Per-channel\\mean} &
        $\|\operatorname{Mean}_l(\tilde{V}_s^t)-\operatorname{Mean}_l(V_s^t)\|_{\infty}$ &
        $3.58\times10^{-7}$ \\
        \midrule
        \shortstack[c]{Per-channel\\variance} &
        $\|\operatorname{Var}_l(\tilde{V}_s^t)-\operatorname{Var}_l(V_s^t)\|_{\infty}$ &
        $0$ \\
        \midrule
        Gram matrix &
        $\displaystyle
        \frac{{\|\tilde{V}_s^t}^{\top}\tilde{V}_s^t-{V_s^t}^{\top}V_s^t\|_F}
        {\|{V_s^t}^{\top}V_s^t\|_F}$ &
        $1.72\times10^{-7}$ \\

        \bottomrule
    \end{tabular}
\end{table}

\paragraph{Why does this experiment identify key--value coupling as a key factor?}
Joint shuffling moves each key together with its original value, so the native key--value correspondence is preserved despite the change in spatial position. 
Therefore, the same semantic information can still be retrieved, leading to nearly unchanged probing performance. 
In contrast, value-only shuffling keeps the keys and attention weights unchanged but assigns each key a value from another position, thereby breaking the native key--value correspondence.
The resulting performance drop indicates that semantic information mainly depends on the coupling between corresponding keys and values, rather than on their spatial positions or the value distribution alone.
\begin{table*}[t]
\centering
\caption{Effect of value-token distribution preservation and K--V decoupling. Best results are highlighted in \textbf{\textcolor{red!80!black}{red bold}}.}
\label{tab:distribution_importance}
\small
\setlength{\tabcolsep}{3.0pt}
\renewcommand{\arraystretch}{0.80}
\begin{tabular}{c c c c c c c c c}
\toprule
\multirow{2}{*}{Setting}
& Preserve empirical
& Break K--V
& \multirow{2}{*}{ArtFID $\downarrow$}
& \multirow{2}{*}{FID $\downarrow$}
& \multirow{2}{*}{Gram $\downarrow$}
& \multirow{2}{*}{LPIPS $\downarrow$}
& \multirow{2}{*}{CFSD $\downarrow$}
& \multirow{2}{*}{DINOv2 $\uparrow$} \\
& distribution
& correspondence
& & & & & & \\
\midrule

No shuffle
& $\checkmark$ & $\times$
& 24.700 & 14.763 & 1.903 & 0.5670 & 0.2755 & 0.4987 \\

Matched Gaussian
& $\times$ & $\checkmark$
& 25.522 & 15.824 & 1.628 & 0.5170 & 0.2441 & 0.5664 \\

Block shuffle ($3\times3$)
& $\checkmark$ & $\checkmark^{\dagger}$
& 23.544 & 14.858 & \textbf{\textcolor{red!80!black}{1.613}} & 0.4847 & 0.2435 & 0.5676 \\

\rowcolor{blue!8}
\textbf{SuRe}
& $\checkmark$ & $\checkmark$
& \textbf{\textcolor{red!80!black}{21.714}}
& \textbf{\textcolor{red!80!black}{13.833}}
& 1.621
& \textbf{\textcolor{red!80!black}{0.4620}}
& \textbf{\textcolor{red!80!black}{0.2049}}
& \textbf{\textcolor{red!80!black}{0.6459}} \\
\bottomrule
\end{tabular}

\vspace{1mm}
\footnotesize $\dagger$ Block shuffle breaks correspondence only across blocks while retaining within-block correspondence.
\end{table*}

\section{Value-Distribution Verification}
\label{sec:value_distribution_verification}
V-Shuffle is an exact permutation of the spatial value tokens, and thus its empirical value distribution is preserved by construction.
To additionally verify the implementation, we save the original values $V_s^t$ and their shuffled counterparts $\tilde{V}_s^t=P_l^tV_s^t$ at each timestep and every selected self-attention layer.
We evaluate one representative content--style pair using SD1.5 with 200 sampling steps and six selected attention layers (layers 10--15), resulting in 1,200 layer--timestep pairs. For each pair, we compare (i) the per-attention-head, per-channel mean and variance over spatial tokens, and (ii) the uncentered token Gram matrix $V^\top V$ between the original and shuffled values.
As shown in Tab.~\ref{tab:value_distribution_verification}, the variance is exactly preserved, while the maximum mean error and relative Gram-matrix error are only $3.58\times10^{-7}$ and $1.72\times10^{-7}$, respectively. 
These negligible differences are due solely to floating-point reduction order, confirming that V-Shuffle preserves value-token statistics while only changing their spatial assignment.

\paragraph{Why is preserving the empirical value distribution important?}
Preserving the empirical distribution of values alone is not sufficient, since the native setting retains both the original empirical distribution and their key--value correspondence. 
Conversely, disrupting the correspondence by directly altering the value empirical distribution may degrade style fidelity.
To disentangle these two factors, we compare four variants under the same PRM protocol, changing only the shuffled-value branch. ``No shuffle'' keeps the native values, Matched Gaussian replaces them with samples having the same mean and variance as $V_s^t$, Block shuffle preserves the original token set but only permutes local $3\times3$ blocks, and SuRe applies the full V-Shuffle to permute all value tokens along the spatial dimension, thereby exactly preserving the empirical distribution while completely breaking the native key–value correspondence.

\begin{figure}[t]
    \centering
    \includegraphics[width=\linewidth]{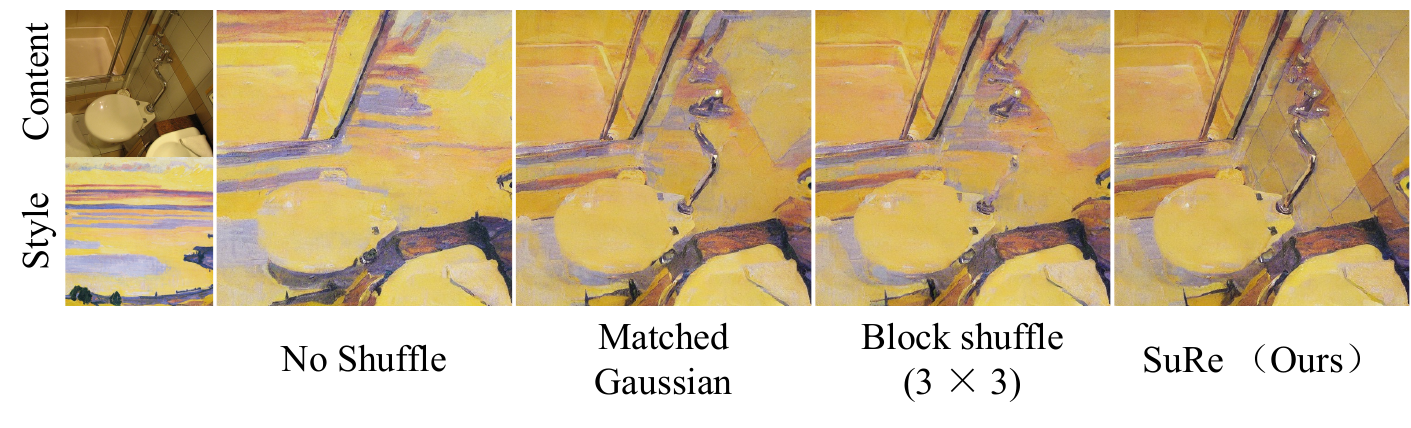}
    \caption{Qualitative comparison of alternative value transformations. SuRe best preserves the content structure while transferring the reference colors and textures.}
\label{fig:distribution_importance}
\end{figure}

As shown in Tab.~\ref{tab:distribution_importance}, SuRe performs best across ArtFID, FID, LPIPS, CFSD, and DINOv2. Fig.~\ref{fig:distribution_importance} further shows that SuRe better preserves the content structure while maintaining the target colors and textures. Together, these results indicate that effective stylization benefits from simultaneously preserving the original value  empirical distribution and sufficiently disrupting the native spatial key--value correspondence.

\section{Attention Distribution under V-Shuffle}
\label{sec:attention_output_change}

In Sec.~\ref{sec:vshuffle}, we prove that V-Shuffle preserves the empirical distribution of value tokens.
Here, we further analyze how V-Shuffle affects the distribution of the resulting attention outputs.
For clarity, we consider one attention head and omit the timestep and style superscripts. The native attention operation is defined as
\begin{equation}
\mathcal{A}
=
\operatorname{Softmax}
\left(
\frac{Q_c K_s^{\top}}{\sqrt{d}}
\right),
\mathcal{O}
=
\mathcal{A} V_s,
\label{eq:native_attention_output}
\end{equation}
where $\mathcal{A}\in\mathbb{R}^{n\times l}$ denotes the attention-weight matrix and $\mathcal{O} \in\mathbb{R}^{n\times d}$ denotes the attention output.

Since V-Shuffle keeps $Q_c$ and $K_s$ unchanged, the attention-weight matrix $\mathcal{A}$ remains fixed. V-Shuffle only replaces the native value $V_s$ with $\tilde{V}_s=P_l V_s$, where $P_l\in\mathbb{R}^{l\times l}$ is a permutation matrix.
Therefore, the shuffled attention output becomes
\begin{equation}
\tilde{\mathcal{O}}
=
\mathcal{A}\tilde{V}_s
=
\mathcal{A}P_lV_s.
\label{eq:shuffled_attention_output}
\end{equation}
For the $i$-th content query, the native output and the shuffled output can be written as
\begin{equation}
\mathcal{O}_i
=
\sum_{j=1}^{l}
\mathcal{A}_{ij} V_s^j,
\quad
\tilde{\mathcal{O}}_i
=
\sum_{j=1}^{l}
\mathcal{A}_{ij} V_s^{\pi(j)},
\end{equation}
where $\pi(\cdot)$ denotes the permutation induced by $P_l$.
By re-indexing the summation, $\tilde{\mathcal{O}}_i$ can equivalently be written as
\begin{equation}
\tilde{\mathcal{O}}_i
=
\sum_{j=1}^{l}
\mathcal{A}_{i,\pi^{-1}(j)} V_s^j.
\label{eq:shuffled_query_output_2}
\end{equation}
Comparing $\mathcal{O}_i$ and $\tilde{\mathcal{O}}_i$, we observe that
the value token $V_s^j$ is originally weighted by
$\mathcal{A}_{ij}$, whereas after V-Shuffle it is weighted by
$\mathcal{A}_{i,\pi^{-1}(j)}$.
Thus, V-Shuffle preserves the set of value tokens but changes
their correspondence with the attention weights.

\begin{figure*}
    \centering
    \includegraphics[width=\textwidth]{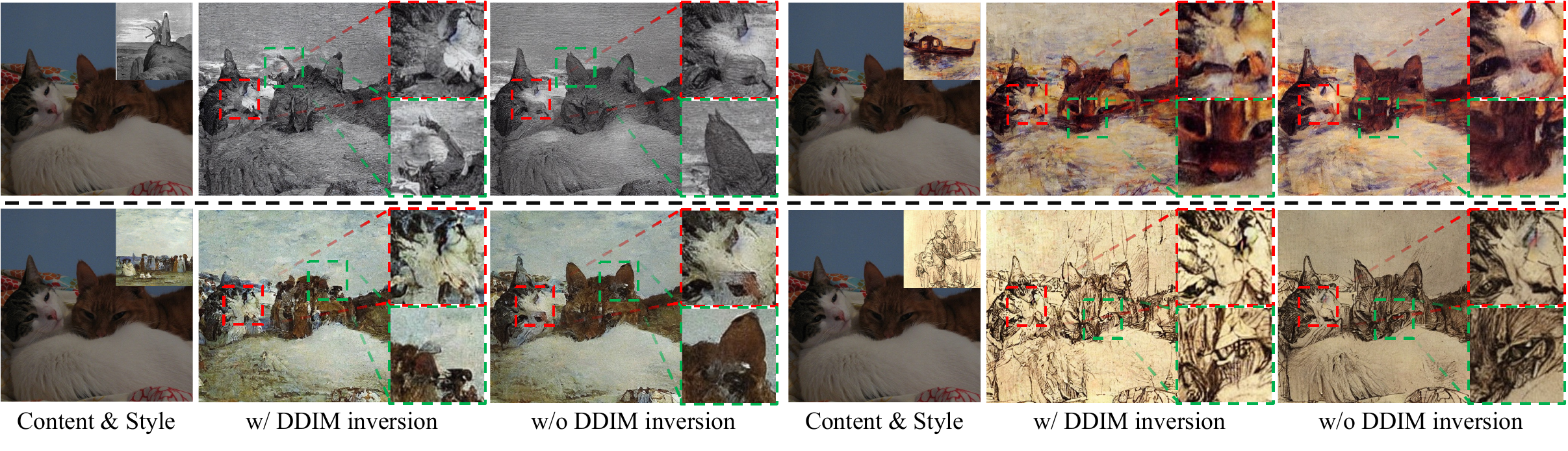}
    \caption{Qualitative effect of DDIM inversion. For each content--style pair, fixed-latent stylization without DDIM inversion better preserves the original content structure and follows the reference appearance. Dashed boxes highlight representative differences.}
    \label{fig:fixed_latent}
\end{figure*}

\begin{table*}[t]
    \centering
    \caption{Effect of DDIM inversion. Best results are highlighted in \textbf{\textcolor{red!80!black}{red}}.}
    \label{tab:fixed_latent}
    \small
    \setlength{\tabcolsep}{5pt}
    \renewcommand{\arraystretch}{0.88}
    \begin{tabular}{l c c c c c c}
        \toprule
        & \multicolumn{3}{c}{Style similarity} & \multicolumn{3}{c}{Content similarity} \\
        \cmidrule(lr){2-4}\cmidrule(lr){5-7}
        Setting & ArtFID $\downarrow$ & FID $\downarrow$ & Gram $\downarrow$ & LPIPS $\downarrow$ & CFSD $\downarrow$ & DINOv2 $\uparrow$ \\
        \midrule
        w/ DDIM inversion & 26.826 & 15.336 & 2.048 & 0.6421 & 0.3211 & 0.3837 \\
        \textbf{w/o DDIM inversion} & \textbf{\textcolor{red!80!black}{21.714}} & \textbf{\textcolor{red!80!black}{13.833}} & \textbf{\textcolor{red!80!black}{1.621}} & \textbf{\textcolor{red!80!black}{0.4620}} & \textbf{\textcolor{red!80!black}{0.2049}} & \textbf{\textcolor{red!80!black}{0.6459}} \\
        \bottomrule
    \end{tabular}
\end{table*}

We now consider the empirical distributions of the attention
output tokens,
\begin{equation}
\widehat{\mu}_{\mathcal{O}}
=
\frac{1}{n}
\sum_{i=1}^{n}
\delta_{\mathcal{O}_i},
\quad
\widehat{\mu}_{\tilde{\mathcal{O}}}
=
\frac{1}{n}
\sum_{i=1}^{n}
\delta_{\tilde{\mathcal{O}}_i},
\label{eq:attention_output_distribution}
\end{equation}
where $\delta_x$ denotes the Dirac measure centered at $x$.
If V-Shuffle preserved the empirical distribution of the attention
outputs, then
\begin{equation}
\widehat{\mu}_{\tilde{\mathcal{O}}}
=
\widehat{\mu}_{\mathcal{O}}.
\label{eq:distribution_equal}
\end{equation}
For two finite empirical distributions with the same number of samples, Eq.~\ref{eq:distribution_equal} implies that the output tokens can differ only by a permutation of their indices.
Therefore, there must exist an
$n\times n$ permutation matrix $R$ such that
\begin{equation}
\tilde{\mathcal{O}}
=
R\mathcal{O}.
\label{eq:output_permutation}
\end{equation}
Substituting $\mathcal{O}=\mathcal{A}V_s$ and $\tilde{\mathcal{O}}=\mathcal{A}P_lV_s$ into Eq.~\ref{eq:output_permutation} gives
\begin{equation}
\mathcal{A}P_lV_s
=
R\mathcal{A}V_s.
\label{eq:distribution_condition_1}
\end{equation}
For general value features $V_s$, this requires
\begin{equation}
\mathcal{A}P_l
=
R\mathcal{A}.
\label{eq:distribution_condition_2}
\end{equation}
However, the left-hand side of Eq.~\ref{eq:distribution_condition_2} permutes the columns of $\mathcal{A}$, while the right-hand side permutes its rows.
The columns of $\mathcal{A}$ correspond to different style keys, whereas its rows correspond to different content queries.
Therefore, Eq.~\ref{eq:distribution_condition_2} requires a special symmetry between the content-query and style-key dimensions, which is not generally satisfied by the attention matrix. Consequently, for a nontrivial permutation $P_l$, we generally have
\begin{equation}
\widehat{\mu}_{\tilde{\mathcal{O}}}
\neq
\widehat{\mu}_{\mathcal{O}}.
\label{eq:distribution_changed}
\end{equation}

Overall, although V-Shuffle preserves the empirical distribution of
the value tokens, it generally changes the empirical distribution of the resulting attention outputs.
The reason is that V-Shuffle preserves the values themselves while
disrupting their native correspondence with the attention weights
determined by $Q_c$ and $K_s$.
As a result, semantic information associated with the native
key--value correspondence can no longer be consistently retrieved by the corresponding content queries, thereby mitigating semantic leakage.

\begin{table}[t]
    \centering
    \caption{Hyperparameters for SuRe across backbones.}
    \label{tab:backbone_hyperparameters}
    \small
    \setlength{\tabcolsep}{1.8pt}
    \renewcommand{\arraystretch}{0.86}
    \begin{tabular}{l c c c c c }
        \toprule
        Backbone & Resolution & $T$ & $\beta$ & LR & Layers \\
        \midrule
        SD1.5~\cite{LDM} & $512\times512$ & 200 & 0.24 & 0.05 & 10--15 \\
        SDXL~\cite{SDXL} & $1024\times1024$ & 200 & 0.30 & 0.02 & 50--56 \\
        SD3~\cite{SD3} & $512\times512$ & 200 & 0.22 & 0.015 & 17--22 \\
        FLUX~\cite{FLUX} & $512\times512$ & 200 & 0.25 & 0.02 & 50--56 \\
        \bottomrule
    \end{tabular}
\end{table}

\section{Analysis of Fixed-Latent Stylization}
\label{sec:fixed_latent_analysis}
Several diffusion attention-based stylization methods~\cite{StyleID,StyleGallery,cocodiff}, obtain image latents through DDIM inversion~\cite{DDIM}. Given the VAE-encoded style and content latents $z_s^0$ and $z_c^0$, this procedure constructs the trajectories $\{z_s^t\}_{t=0}^{T}$ and $\{z_c^t\}_{t=0}^{T}$. The diffusion model then extracts the corresponding attention features at each timestep as
\begin{equation}
\begin{aligned}
\{Q_s^t,K_s^t,V_s^t\} &= \epsilon_\theta(z_s^t,t,\emptyset),\\
\{Q_c^t,K_c^t,V_c^t\} &= \epsilon_\theta(z_c^t,t,\emptyset).
\end{aligned}
\label{eq:ddim_feature_extraction}
\end{equation}
In contrast, SuRe keeps $z_s^0$ and $z_c^0$ fixed throughout stylization, and changes only the timestep supplied to the diffusion model:
\begin{equation}
\begin{aligned}
\{Q_s^t,K_s^t,V_s^t\} &= \epsilon_\theta(z_s^0,t,\emptyset),\\
\{Q_c^t,K_c^t,V_c^t\} &= \epsilon_\theta(z_c^0,t,\emptyset).
\end{aligned}
\label{eq:fixed_latent_feature_extraction}
\end{equation}

We conduct this ablation on SD1.5. 
Tab.~\ref{tab:fixed_latent} compares the two feature-extraction schemes. 
Avoiding DDIM inversion improves all reported metrics. 
In particular, it reduces LPIPS from 0.6421 to 0.4620 and CFSD from 0.3211 to 0.2049, while increasing the DINOv2 score from 0.3837 to 0.6459. It also improves ArtFID, FID, and Gram loss, indicating that fixed latent inputs preserve both content structure and style fidelity more effectively in this setting. 
These results suggest that repeatedly noising the image latents can interfere with the pretrained diffusion model's extraction of content and style features for attention matching. Fig.~\ref{fig:fixed_latent} provides a qualitative comparison. 
DDIM inversion introduces visible content distortions, whereas the fixed-latent setting preserves the content layout while reproducing the target colors and textures. This observation is consistent with the quantitative comparison in Tab.~\ref{tab:fixed_latent}.

\begin{table}[t]
    \centering
    \caption{Efficiency comparison across evaluated methods. Peak memory is reported in GB and inference time in seconds.}
    \label{tab:efficiency}
    \small
    \setlength{\tabcolsep}{1.2pt}
    \renewcommand{\arraystretch}{0.8}
    \begin{tabular}{c l c r r}
        \toprule
        Category & Method & Backbone & Memory $\downarrow$ & Time $\downarrow$ \\
        \midrule
        \multirow{7}{*}{\shortstack[c]{Feed-forward\\methods}}
        & AdaIN~\cite{AdaIN} & CNN & 2.09 & 0.58 \\
        & MAST~\cite{MAST} & CNN & 1.11 & 2.68 \\
        & EFDM~\cite{EFDM} & CNN & 2.09 & 0.53 \\
        & CAST~\cite{CAST} & CNN & 33.09 & 6.04 \\
        & AesPA-Net~\cite{Aespa-net} & CNN & 3.19 & 4.01 \\
        & StyTR$^2$~\cite{StyTR2} & Transformer & 4.34 & 0.82 \\
        & AdaAttN~\cite{AdaAttN} & Transformer & 33.34 & 9.13 \\
        \midrule
        \multirow{5}{*}{\shortstack[c]{Diffusion \\adaptation\\methods}}
        & AnyStyle~\cite{AnyStyle} & Diffusion & 33.17 & 25.38 \\
        & OmniStyle~\cite{omnistyle} & Flow & 35.65 & 17.46 \\
        & K-LoRA~\cite{K-lora} & Diffusion & 14.64 & 32.57 \\
        & InstantStyle~\cite{instantstyle} & Diffusion & 16.02 & 6.32 \\
        & InST~\cite{InST} & Diffusion & 11.05 & 4.91 \\
        \midrule
        \multirow{10}{*}{\shortstack[c]{Diffusion\\attention\\methods}}
        & CoCoDiff~\cite{cocodiff} & Diffusion & 14.06 & 19.93 \\
        & ACID-Style~\cite{ACID-Style} & Diffusion & 8.07 & 2.06 \\
        & StyleGallery~\cite{StyleGallery} & Diffusion & 4.00 & 30.01 \\
        & StyleSSP~\cite{Stylessp} & Diffusion & 20.27 & 20.11 \\
        & AD~\cite{AD} & Diffusion & 4.34 & 41.79 \\
        & StyleID~\cite{StyleID} & Diffusion & 19.38 & 17.80 \\
        \cmidrule(lr){2-5}
        & \cellcolor{blue!8}\textbf{Ours (SD1.5)} & \cellcolor{blue!8}Diffusion & \cellcolor{blue!8}3.39 & \cellcolor{blue!8}45.41 \\
        & \cellcolor{blue!8}\textbf{Ours (SDXL)} & \cellcolor{blue!8}Diffusion & \cellcolor{blue!8}12.08 & \cellcolor{blue!8}113.76 \\
        & \cellcolor{blue!8}\textbf{Ours (SD3)} & \cellcolor{blue!8}Flow & \cellcolor{blue!8}16.43 & \cellcolor{blue!8}45.43 \\
        & \cellcolor{blue!8}\textbf{Ours (FLUX)} & \cellcolor{blue!8}Flow & \cellcolor{blue!8}31.46 & \cellcolor{blue!8}91.85 \\
        \bottomrule
    \end{tabular}
\end{table}

\section{Algorithm of SuRe}
\label{sec:algorithm_supp}

\begin{algorithm}[t]
\caption{Algorithmic pseudocode of SuRe}
\label{alg:sure}
\small
\begin{algorithmic}[1]
\Require Content image $I_c$, style image $I_s$, VAE encoder $\mathcal{E}$, VAE decoder $\mathcal{D}$, diffusion model $\epsilon_\theta$, sampling timesteps $T$, number of permutation matrices $m$, and content-loss weight $\beta$
\Ensure Stylized image $I_{cs}$
\State $z^0_c\gets\mathcal{E}(I_c)$; $z^0_s\gets\mathcal{E}(I_s)$
\State $z^T_{cs}\gets z^0_c$
\For{$t=T,T-1,\ldots,1$}
    \State $\{Q_c^t,K_c^t,V_c^t\}\gets\epsilon_\theta(z_0^c,t,\emptyset)$
    \State $\{Q_s^t,K_s^t,V_s^t\}\gets\epsilon_\theta(z_0^s,t,\emptyset)$
    \State $\mathcal{O}_{\mathrm{nat}}\gets\textbf{A}(Q_c^t,K_s^t,V_s^t)$
    \State $\alpha_t\gets\frac{1}{2}\left[1+\cos\left(\frac{\pi t}{T}\right)\right]$ \Comment{cosine schedule}
    \For{$r=1,\ldots,m$}
        \State Sample a permutation matrix $P_l^{t,(r)}\in\{0,1\}^{l\times l}$
        \State $\tilde V_s^{t,(r)}\gets P_l^{t,(r)}V_s^t$ \Comment{V-Shuffle}
        \State $\mathcal{O}_{\mathrm{sh}}^{(r)}\gets\textbf{A}(Q_c^t,K_s^t,\tilde V_s^{t,(r)})$
        \EndFor
    \State $\{Q_{cs}^t,K_{cs}^t,V_{cs}^t\}\gets\epsilon_\theta(z^t_{cs},t,\emptyset)$
    \State $\mathcal{O}_{cs}\gets\textbf{A}(Q_{cs}^t,K_{cs}^t,V_{cs}^t)$
    \State $\mathcal{L}_s\gets\alpha_t\|\mathcal{O}_{cs}-\mathcal{O}_{\mathrm{nat}}\|_1$ \Comment{PRM loss}
    \Statex \hspace{\algorithmicindent}\hspace{\algorithmicindent}$+(1-\alpha_t)\frac{1}{m}\sum_{r=1}^{m}\|\mathcal{O}_{cs}-\mathcal{O}_{\mathrm{sh}}^{(r)}\|_1$
    \State $\mathcal{L}_c\gets\|Q_{cs}^t-Q_c^t\|_1$
    \State $\mathcal{L}_{\mathrm{total}}\gets\mathcal{L}_s+\beta\mathcal{L}_c$
    \State $z^{t-1}_{cs} \gets \arg\min \mathcal{L}_{total}$
\EndFor
\State \Return $I_{cs}\gets\mathcal{D}(z^0_{cs})$
\end{algorithmic}
\end{algorithm}

Alg.~\ref{alg:sure} summarizes the complete sampling procedure. At each timestep, V-Shuffle independently samples $m$ permutation matrices and applies each of them only to the spatial dimension of the style values. PRM then uses the cosine weight $\alpha_t$ to transition from shuffled attention targets to the native attention target.

\begin{figure*}[t]
    \centering
    \begin{minipage}[t]{0.48\textwidth}
        \centering
        \includegraphics[width=\linewidth]{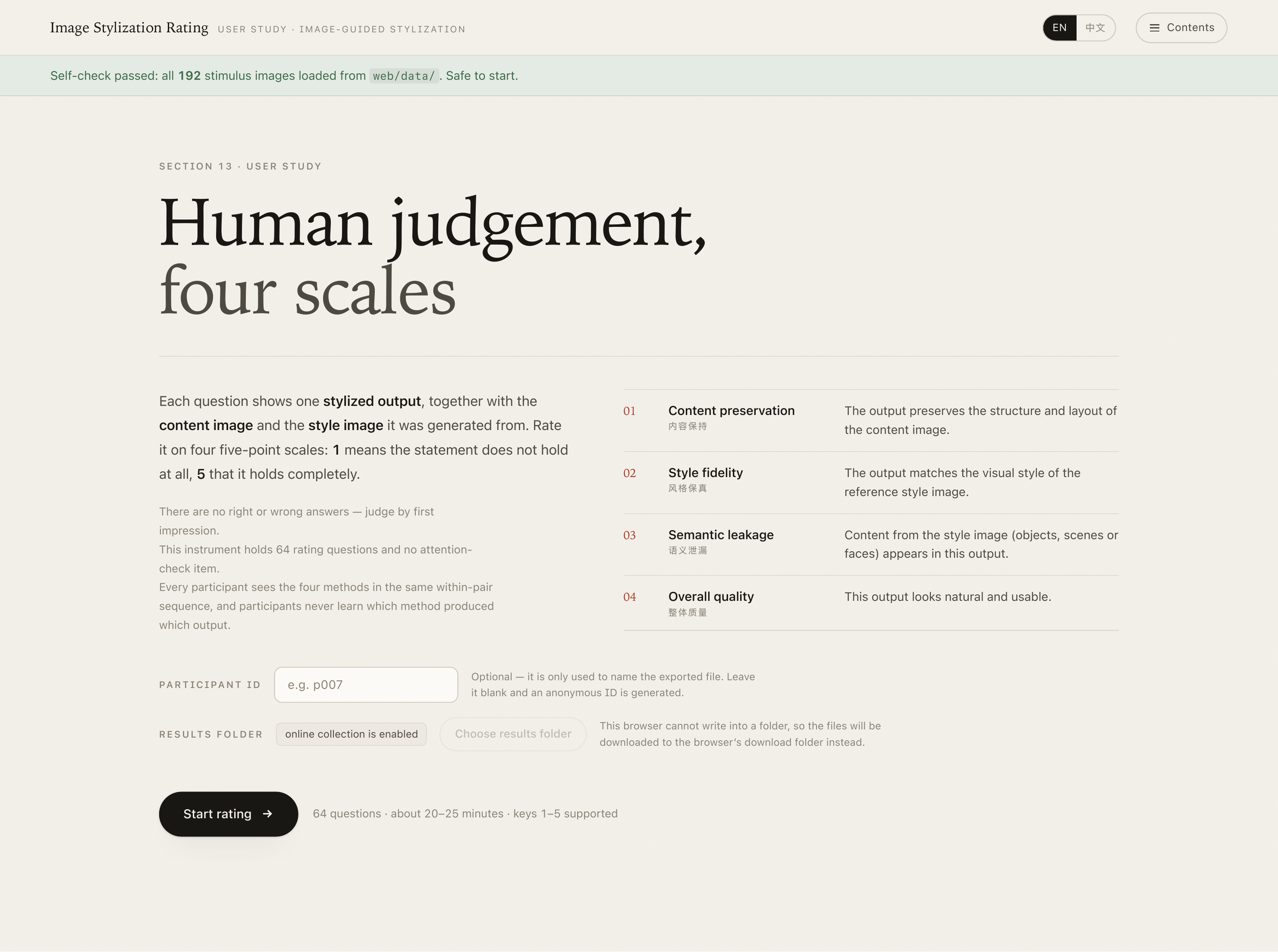}
        \caption{Introduction page of the user study.}
        \label{fig:user_study_start}
    \end{minipage}
    \hfill
    \begin{minipage}[t]{0.50\textwidth}
        \centering
        \includegraphics[width=\linewidth]{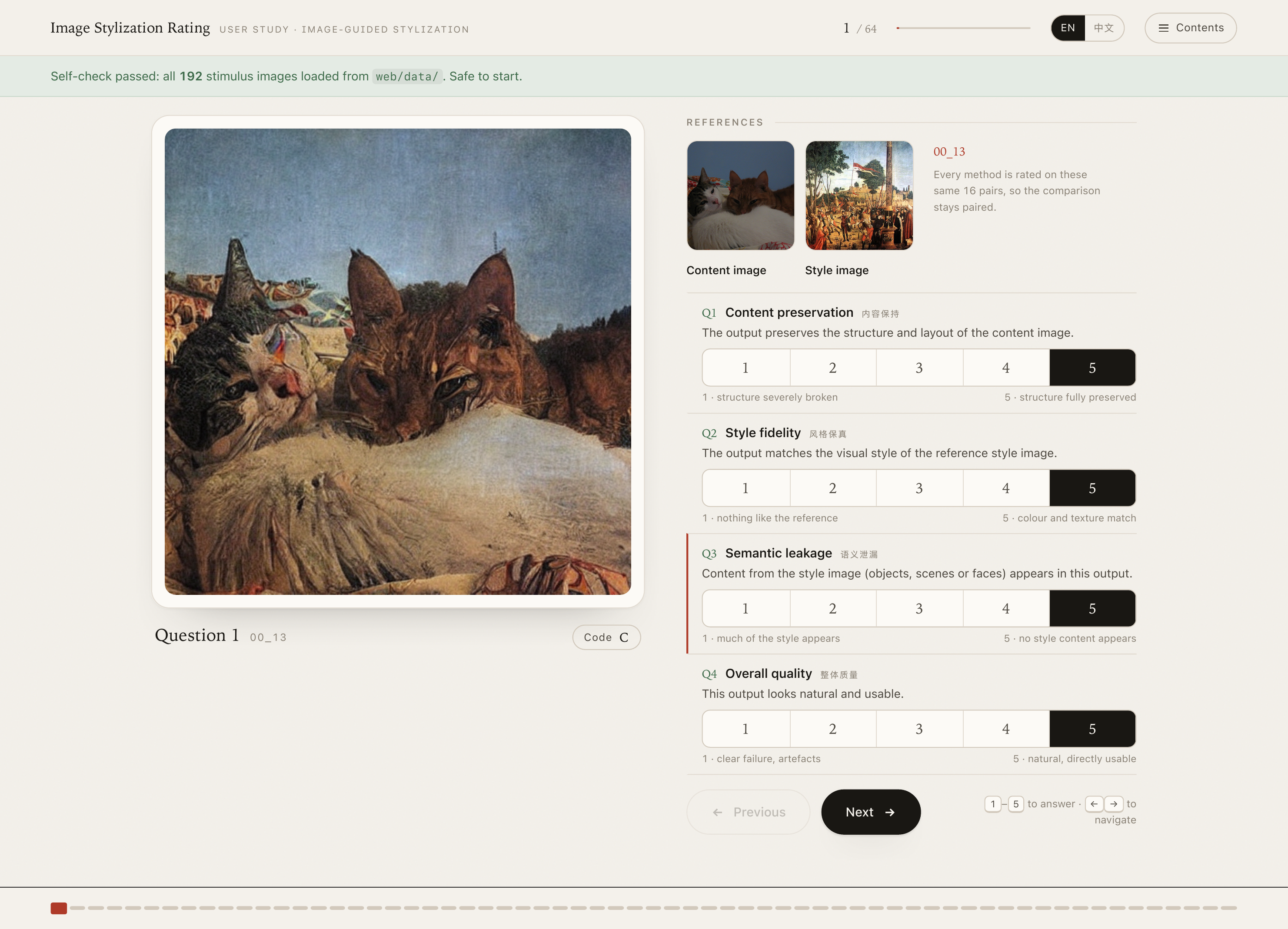}
        \caption{Evaluation interface of the user study.}
        \label{fig:example1}
    \end{minipage}
\end{figure*}

\section{Hyperparameters for Different Backbones}
\label{sec:backbone_hyperparameters}
Tab.~\ref{tab:backbone_hyperparameters} lists the hyperparameters used for all backbones.
We use 200 sampling steps for all backbones and optimize the stylized latent with one Adam update per timestep.
The content-loss weight $\beta$, learning rate, and selected self-attention layers are adjusted according to the backbone architecture.
Additionally, we run SDXL at its native $1024\times1024$ resolution because reducing the resolution to $512\times512$ introduced substantial visual artifacts.  

\section{Efficiency comparison}
\label{sec: efficiency comparison}
Tab.~\ref{tab:efficiency} compares peak GPU memory and inference time across the evaluated methods. 
Conventional feed-forward methods generally achieve shorter inference times, while diffusion-based methods show greater variation due to differences in backbones and sampling or optimization procedures. 
Among the attention-based diffusion methods, SuRe with SD1.5 requires 3.39~GB of peak memory, lower than most baselines in this group, while its 45.41~s inference time mainly results from iterative latent optimization and the use of five shuffled attention targets at each timestep. 
The SDXL, SD3, and FLUX.1-dev variants require more memory with larger backbones. 
Since these methods differ in model families and inference procedures, the results primarily reflect their computational resource requirements.

\begin{figure}[t]
    \centering
    \includegraphics[width=0.95\linewidth]{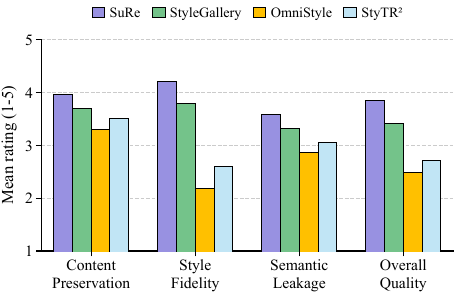}
    \caption{Mean user-study ratings across 20 blinded participants for content preservation, style fidelity, semantic leakage control, and overall quality. Ratings use a 1--5 Likert scale; higher scores indicate better performance.}
    \label{fig:rating}
\end{figure}

\section{User Study}
\label{sec:user_study}
To evaluate the qualitative quality of stylized images, we conduct a user study comparing our method with representative methods (StyTR$^2$~\cite{StyTR2}, StyleGallery~\cite{StyleGallery}, and OmniStyle~\cite{omnistyle}) that exhibit strong overall quantitative performance in their respective categories.
We prepare 64 samples, each consisting of a content image, a style image, and a stylized image (192 images in total).
Each sample is rated on four criteria using a five-point
scale (1: worst; 5: best):

\begin{itemize}
    \item \textbf{Content Preservation:}
    The extent to which the stylized image preserves
    the content structure and layout.
    
    \textit{(1: severe distortion; 5: fully preserved).}

    \item \textbf{Style Fidelity:}
    The extent to which the stylized image reflects
    the reference style.
    
    \textit{(1: inconsistent style; 5: highly consistent
    style).}

    \item \textbf{Semantic Leakage:}
    The extent to which semantic content from the
    style image appears in the stylized image.
    
    \textit{(1: severe leakage; 5: no leakage).}

    \item \textbf{Overall Quality:}
    The naturalness and usability of the stylized image.
    
    \textit{(1: obvious artifacts; 5: natural and usable).}
\end{itemize}

We recruit 20 human participants for the evaluation, collecting 1,280 responses in total.
The web interface of our user study is illustrated in Figs.~\ref{fig:user_study_start}--\ref{fig:example1}. 

As shown in Fig.~\ref{fig:rating}, SuRe achieves the highest mean ratings across all four criteria, demonstrating its superior perceptual quality. Specifically, SuRe obtains a Content Preservation score of 3.97, outperforming all competing methods and indicating its ability to maintain content structure during stylization. For Style Fidelity, SuRe achieves the highest score of 4.2, compared with 3.8 for StyleGallery, 2.2 for OmniStyle, and 2.6 for StyTR$^2$, highlighting its effectiveness in transferring reference styles.

Notably, SuRe also achieves the highest Semantic Leakage score of 3.6, indicating less unwanted semantic transfer from the style image. This suggests that SuRe can better preserve stylistic characteristics while mitigating semantic interference. Furthermore, SuRe achieves an Overall Quality score of approximately 3.9, surpassing the strongest baseline, StyleGallery, by about 0.4 points. These results demonstrate that SuRe achieves a favorable balance among content preservation, style fidelity, and semantic leakage control, leading to improved overall perceptual quality.

\section{Additional Results}
\begin{figure*}[t]
    \centering
    \includegraphics[width=\textwidth]{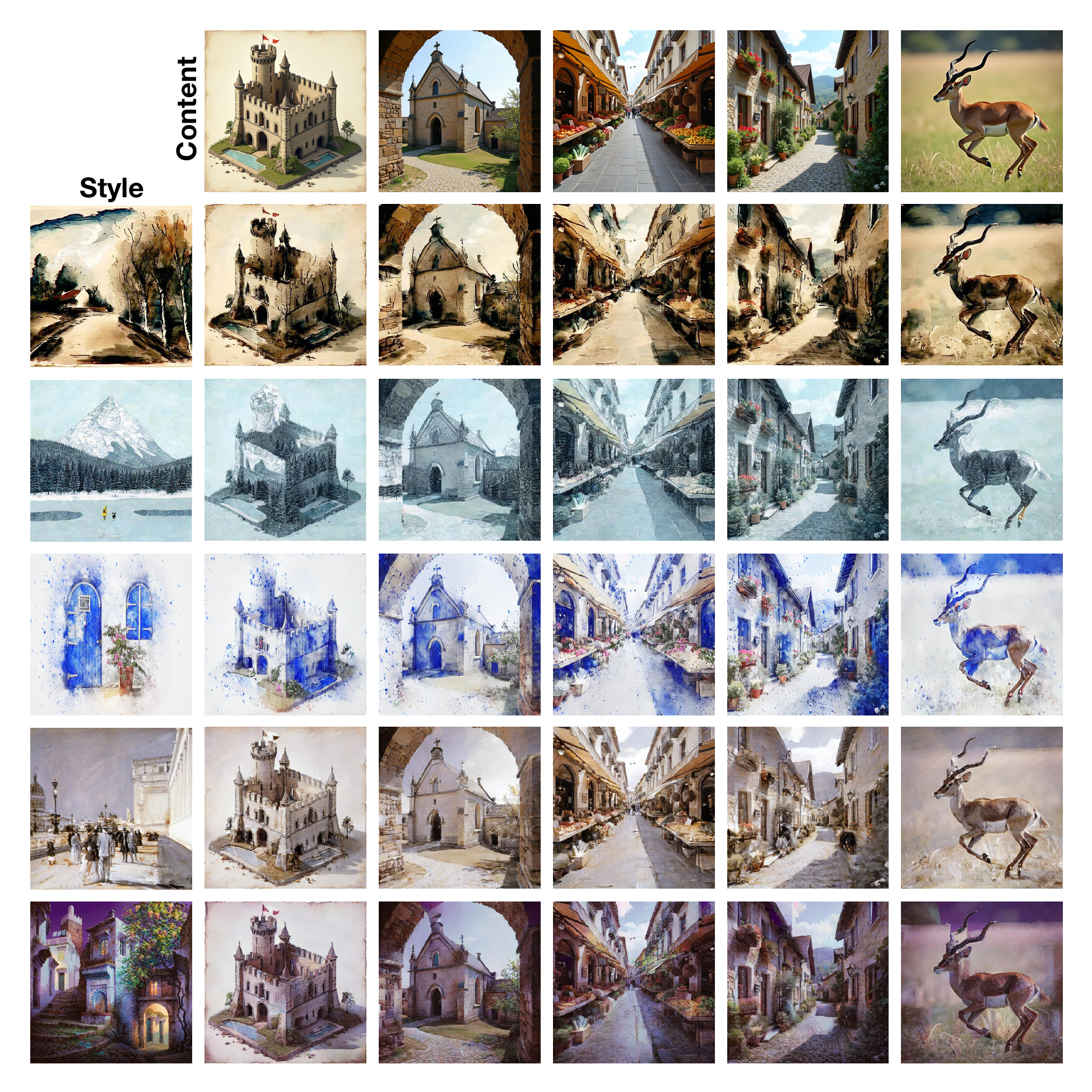} 
    \caption{Additional results of our approach on artistic style transfer.}
    \label{figure 13} 
\end{figure*}

\begin{figure*}[t]
    \centering
    \includegraphics[width=\textwidth]{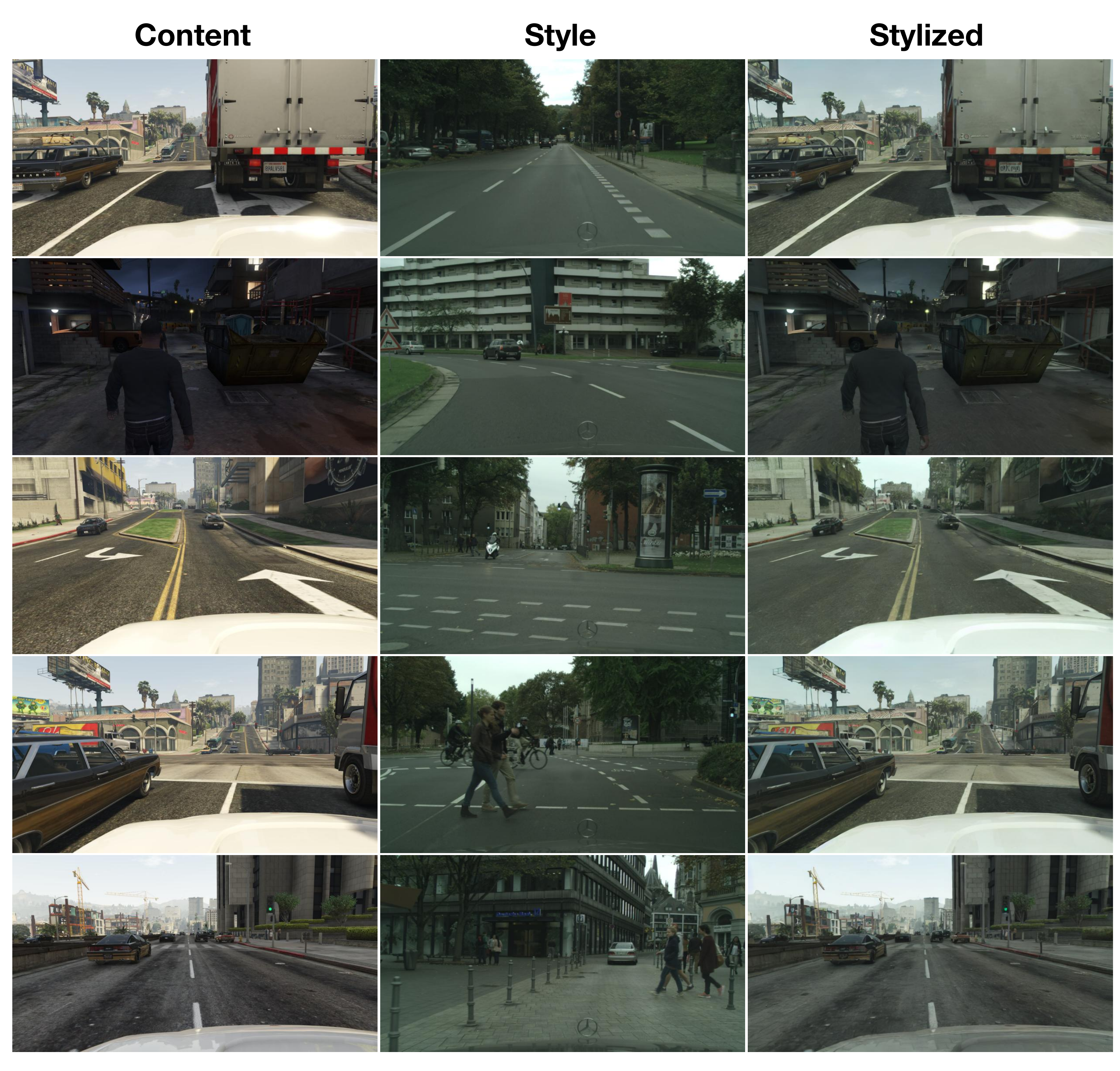} 
    \caption{Additional results of our approach on simulation to real.}
    \label{figure 14} 
\end{figure*}

\begin{figure*}[t]
    \centering
    \includegraphics[width=\textwidth]{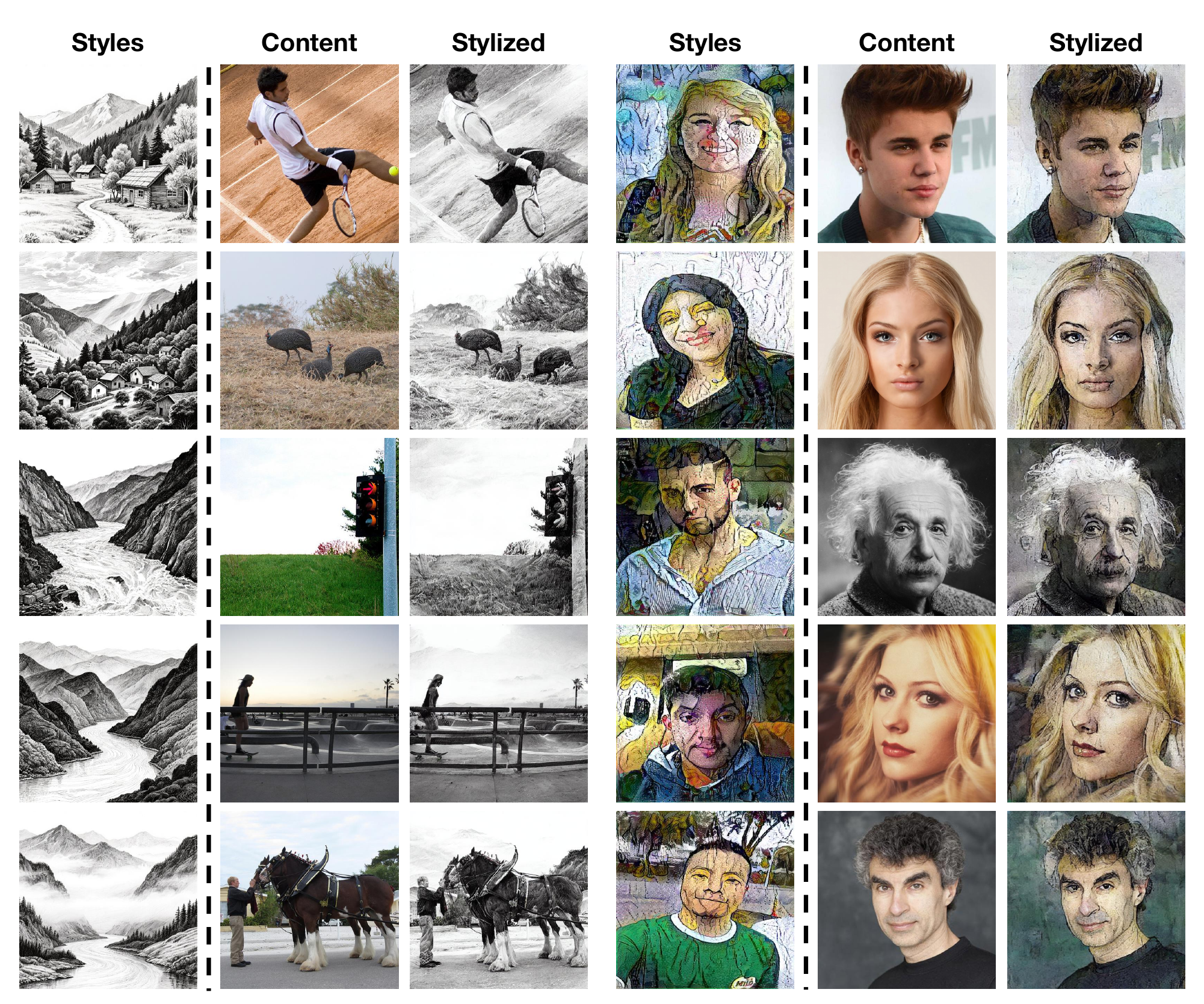} 
    \caption{Additional results of our approach on few-style transfer.}
    \label{figure 15} 
\end{figure*}

Figs.~\ref{figure 13},~\ref{figure 14}, and~\ref{figure 15} provide additional visual results.

\clearpage

\end{document}